\documentclass[sigconf,nonacm]{acmart}

\usepackage{amsmath}

\usepackage{amssymb}
\usepackage{mathtools}
\usepackage{amsthm}
\usepackage{tabularx}

\usepackage{subcaption}  
\usepackage{times}
\usepackage{soul}
\usepackage[utf8]{inputenc}
\usepackage{caption}
\usepackage{graphicx}
\usepackage{dsfont}
\usepackage{enumitem}
\usepackage{booktabs}
\usepackage{algorithm}
\usepackage{algorithmic}
\usepackage[switch]{lineno}
\usepackage{csquotes}
\usepackage{xspace}
\usepackage[dvipsnames]{xcolor}
\usepackage{tcolorbox}
\tcbuselibrary{theorems}

\usepackage[capitalize,noabbrev]{cleveref}

\theoremstyle{plain}
\newtheorem{theorem}{Theorem}[section]

\newtheorem{lemma}[theorem]{Lemma}

\theoremstyle{definition}

\theoremstyle{remark}

\usepackage[textsize=tiny]{todonotes}
\usepackage{microtype}
\usepackage{graphicx}
\usepackage{subcaption}
\usepackage{booktabs} 

\usepackage{hyperref}
\AtBeginDocument{%
  }

\copyrightyear{2026}
\acmYear{2026}
\acmDOI{XXXXXXX.XXXXXXX}

\setcopyright{none}

\acmConference[Conference, AI for Sciences]{Make sure to enter the correct
  conference title from your rights confirmation email}{Washington, DC}

\newcommand{\pname}{\texttt{FOCUS}\xspace}

\setcopyright{none}
\acmConference{}{}{}
\acmBooktitle{}
\acmYear{}
\acmISBN{}
\acmDOI{}

\begin{document}

\title[FOCUS on Contamination]{FOCUS on Contamination: Hydrology-Informed Noise-Aware Learning for Geospatial PFAS Mapping}


\author{Jowaria Khan}
\affiliation{%
  \institution{University of Michigan}
  \city{Ann Arbor}
  \state{Michigan}
  \country{USA}
}
\email{jowaria@umich.edu}

\author{Alexa Friedman}
\affiliation{%
  \institution{Environmental Working Group}
  \city{Washington}
  \state{DC}
  \country{USA}
}
\email{alexa.friedman@ewg.org}

\author{Sydney Evans}
\affiliation{%
  \institution{Environmental Working Group}
  \city{Washington}
  \state{DC}
  \country{USA}
}
\email{sydney.evans@ewg.org}

\author{Rachel Klein}
\affiliation{%
  \institution{University of Michigan}
  \city{Ann Arbor}
  \state{Michigan}
  \country{USA}
}
\email{raklein@umich.edu}

\author{Runzi Wang}
\affiliation{%
  \institution{University of California, Davis}
  \city{Davis}
  \state{California}
  \country{USA}
}
\email{mrrwang@ucdavis.edu}

\author{Katherine E. Manz}
\affiliation{%
  \institution{University of Michigan}
  \city{Ann Arbor}
  \state{Michigan}
  \country{USA}
}
\email{katmanz@umich.edu}

\author{Kaley Beins}
\affiliation{%
  \institution{Environmental Working Group}
  \city{Washington}
  \state{DC}
  \country{USA}
}
\email{kaley.beins@ewg.org}

\author{David Q. Andrews}
\affiliation{%
  \institution{Environmental Working Group}
  \city{Washington}
  \state{DC}
  \country{USA}
}
\email{dandrews@ewg.org}

\author{Elizabeth Bondi-Kelly}
\affiliation{%
  \institution{University of Michigan}
  \city{Ann Arbor}
  \state{Michigan}
  \country{USA}
}
\email{ecbk@umich.edu}

\renewcommand{\shortauthors}{Khan et al.}


\begin{abstract}
Per- and polyfluoroalkyl substances (PFAS) are persistent environmental contaminants with significant public health impacts, yet large-scale monitoring remains severely limited due to the high cost and logistical challenges of field sampling. The lack of samples leads to 
difficulty simulating their spread with physical models and limited scientific understanding of PFAS transport in surface waters. Yet, rich geospatial and satellite-derived data describing land cover, hydrology, and industrial activity are widely available. 
We introduce \pname, a geospatial deep learning framework for PFAS contamination mapping that integrates sparse PFAS observations with large-scale environmental context, including priors derived from hydrological connectivity, land cover, source proximity, and sampling distance. 
These priors are integrated into a principled, noise-aware loss, yielding a robust training objective under sparse labels. Across extensive ablations, robustness analyses, and real-world validation, \pname consistently outperforms baselines including sparse segmentation, Kriging, and pollutant transport simulations, while preserving spatial coherence and scalability over large regions. 
Our results demonstrate how AI can support environmental science by providing screening-level risk maps that prioritize follow-up sampling and help connect potential sources to surface-water contamination patterns in the absence of complete physical models.
\end{abstract}

\ccsdesc[500]{Computing methodologies~Machine learning}
\ccsdesc[300]{Computing methodologies~Learning with noisy labels}
\ccsdesc[300]{Applied computing~Environmental sciences}




\maketitle

\section{Introduction}

\begin{figure}[t]
  \centering
  \includegraphics[width=0.9\columnwidth]{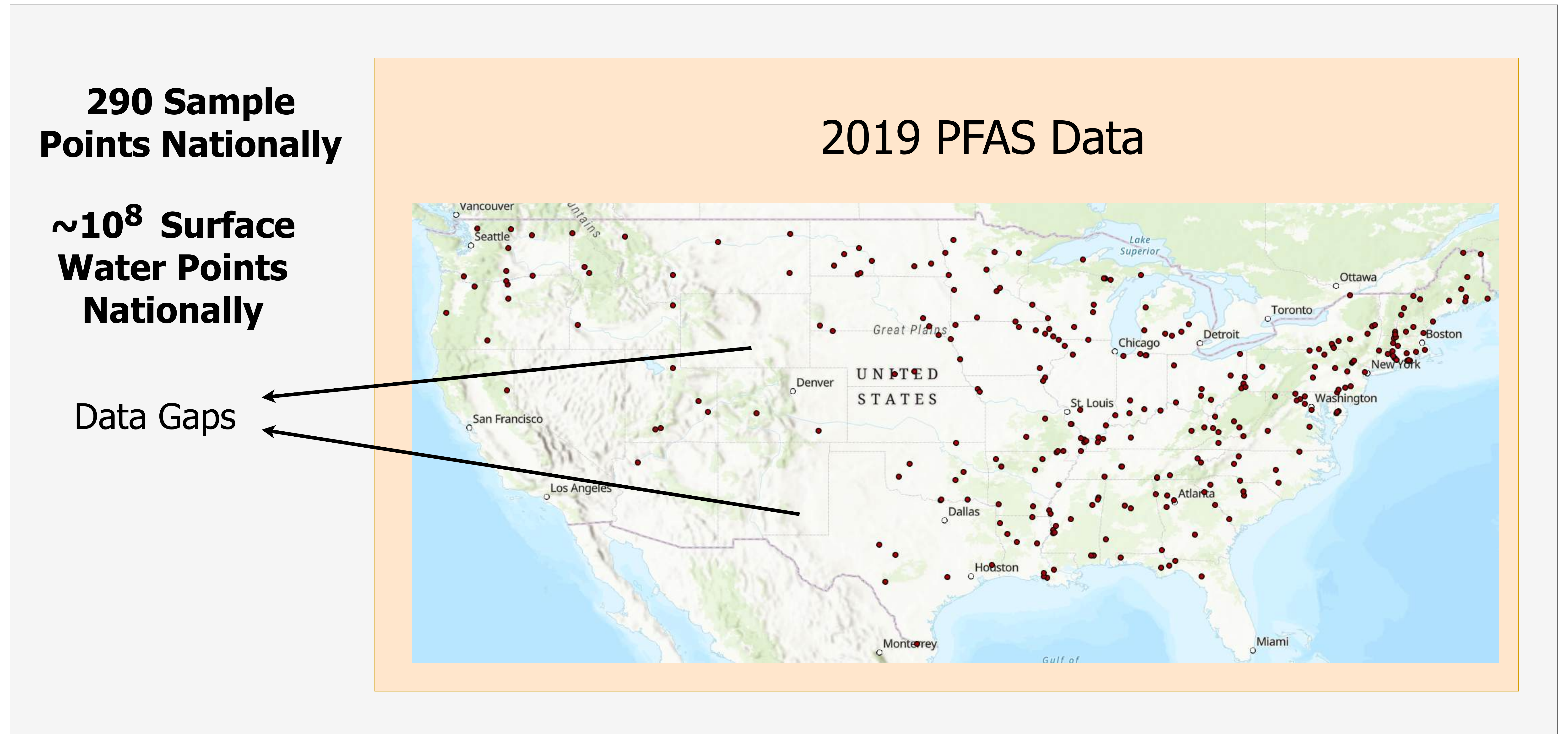}
  \caption{Illustration of limited PFAS contamination data  \protect\cite{USEPA2015}}
  \label{fig:your-image-labell}
\end{figure}

PFAS, or per- and polyfluoroalkyl substances, are persistent “forever chemicals” widely used in industrial and consumer products such as non-stick cookware, waterproof textiles, and firefighting foams \cite{niehs_pfas}. Due to their extreme chemical stability, PFAS resist degradation and are now pervasive across environmental compartments, accumulating in surface and groundwater \cite{Langenbach_Wilson_2021,Schroeder_Bond_Foley_2021}, soils \cite{Crone_Speth_Wahman_Smith_Abulikemu_Kleiner_Pressman_2019}, and living organisms \cite{ewg2023foreverfish}. This widespread environmental presence has direct public health consequences, with 97\% of Americans exhibiting detectable levels in their blood \cite{CDC_2024} and exposure linked to cancers, liver damage, and developmental disorders \cite{Manz_2024}. However, measuring PFAS concentrations remains expensive—often costing hundreds of U.S. dollars per sample—resulting in sparse and unevenly distributed ground-truth data (Fig.~\ref{fig:your-image-labell}). This data scarcity leaves substantial uncertainty in identifying contamination hotspots and prioritizing targeted remediation to best protect public health.



\begin{figure*}[t]
  \centering
  \includegraphics[width=0.85\textwidth]{chals-compressed.pdf}
  \caption{Left: Classical ML techniques aggregate surrounding pixel data to predict contamination at specific points (depicted as x, y coordinates); Right:  DL methods process raster images directly to generate dense PFAS contamination maps in a single pass. }
  \label{fig:test4}
\end{figure*}


Artificial intelligence (AI) presents a promising avenue for addressing complex health and environmental issues, including identifying PFAS hotspots. In various domains, AI has proven effective at automating labor-intensive tasks and synthesizing domain expertise, whether in land cover mapping \cite{Robinson_Ortiz_Malkin_Elias_Peng_Morris_Dilkina_Jojic_2020}, agricultural optimization \cite{Kerner_Nakalembe_Yang_Zvonkov_McWeeny_Tseng_Becker-Reshef_2024}, disease prediction \cite{Bondi-Kelly_Chen_Golden_Behari_Tambe_2023}, or environmental and Earth-system inference tasks \cite{Reichstein_Camps-Valls_Stevens_Jung_Denzler_Carvalhais_Prabhat_2019, Christin_Hervet_Lecomte_2019}. In the domain of PFAS contamination prediction, however, prior work has largely relied on tabular ML models such as random forests \cite{Breiman_2001} and XGBoost \cite{Chen_Guestrin_2016}, which aggregate environmental features around sparse sample points and discard spatial structure. For example, \cite{DeLuca_Mullikin_Brumm_2023} uses a random forest model with features aggregated within a 5 km buffer around each sample (Fig.~\ref{fig:test4}). Building on these efforts, we instead frame PFAS prediction as a geospatial deep learning (DL) task that operates directly on raster inputs, preserving spatial dependencies while reducing the need for manual feature engineering and aggregation.


\noindent Our contributions in this work are summarized as follows:
\begin{itemize}
\item We propose \pname, a principled noise-aware geospatial learning framework comprising a hydrology informed noise-aware objective that combines hydrological priors into a focal-style loss to train robust segmentation models from sparse, noisy PFAS measurements.
\item We formalize PFAS surface-water mapping as learning under \emph{structured, spatially dependent label noise} induced by expanding point measurements to pixel-level supervision.
\item We derive a theoretical result showing that \pname optimizes a valid surrogate objective under asymmetric, pixel-wise label noise, providing a principled justification for the proposed training loss.
\item We construct pixel-wise confidence weights from hydrological flow, industrial discharge proximity, land cover, and sampling distance, linking environmental processes to robust learning.
\item We evaluate on real PFAS field data across the United States and benchmark against geostatistical methods, pollutant transport simulations, and tabular ML baselines, demonstrating improved predictive performance and scalability.
\end{itemize}


\textbf{Domain co-design and motivation.} PFAS mapping poses domain-specific challenges that limit both traditional modeling and off-the-shelf ML: sparse and biased measurements, expensive and uncertain physical simulations, and heavy reliance on manual feature engineering and spatial aggregation. We developed \pname in close collaboration with \textbf{environmental scientists from the Environmental Working Group, academic researchers specializing in water quality models and environmental chemistry, and an advisory group with representatives from other nonprofits and agencies such as the Huron River Watershed Council (HRWC)}, who helped shape both the model inputs (e.g., hydrology, land cover, source proximity) and the learning objective to reflect real scientific uncertainty and regulatory priorities. This co-design process informed our choice to operate directly on geospatial rasters, and to encode physically motivated correctness priors.

\textbf{Reproducibility}. Code, processed datasets, and scripts to reproduce the \pname experiments are available in the accompanying repository \footnote{Code available at \url{https://anonymous.4open.science/r/FOCUS_KDD-D184/README.md}.}.


\section{Related Work}

We now summarize existing research about PFAS, spanning analytical methods from scientific domains, to emerging computational approaches, including AI techniques. We also discuss  AI techniques for geospatial and sparse data.



\noindent\textbf{Current Ways to Measure and Predict PFAS.}
The measurement of PFAS  relies on laboratory-based analysis of water, fish tissue, and human blood samples with techniques such as liquid chromatography-tandem mass spectrometry (LC-MS) \cite{Shoemaker_2020_Method537}.  
PFAS monitoring efforts have  made use of such  measurement techniques and beyond.  
For example, states such as Colorado have implemented water testing, treatment grants, and regulatory measures since 2016 \cite{CDPHE_PFAS_Action_Plan}. In addition, there are community-based water sampling initiatives such as volunteer sampling by the Sierra Club \cite{CommunityScience} and focused efforts by  organizations like the  Ecology Center  \cite{ProtectingCommunitiesFromPFAS}. A recent study further underscores the importance of these efforts, revealing associations between PFAS contamination in drinking water and higher COVID-19 mortality rates in the U.S. \cite{Liddie_Bind_Karra_Sunderland_2024}. 
Resources like the PFAS-Tox Database \cite{Pelch_Reade_Kwiatkowski_Merced-Nieves_Cavalier_Schultz_Wolffe_Varshavsky_2022} also assist   communities in monitoring PFAS. 
While highly accurate and important, laboratory-based  PFAS measurements are costly, time-intensive, and difficult to scale spatially \cite{Doudrick_2024}. 

Various modeling approaches have been explored to predict contaminant distribution. 
Hydrological models such as SWAT \cite{SWAT2023} and MODFLOW \cite{ktorcoletti_2012} simulate pollutant transport using environmental parameters like flow rates, land use, and weather conditions. While effective, these models require extensive data 
and are computationally expensive at large scales \cite{zhi2024deep}. 
%
Kriging, a geostatistical interpolation technique, has been widely used for mapping pollutants, including in soil  \cite{Largueche_2006} and groundwater  \cite{Singh_Verma_2019}. Though useful in data-sparse environments, Kriging relies on simplifying assumptions such as spatial continuity and stationarity \cite{GISGeography_2017} that may not hold for complex contaminants like PFAS, whose transport dynamics and uncertainties are  more intricate (see Section 5). 
Complementing these efforts, Salvatore et al.~\cite{Salvatore_Mok_Garrett_2022} proposed a “presumptive contamination” model that estimates PFAS presence based solely on proximity to known PFAS-using or -producing facilities.
We aim to contribute to monitoring and modeling efforts 
by providing scalable, data-driven PFAS predictions. 

\noindent\textbf{AI Approaches for PFAS Prediction.}
Recent advances in AI and geospatial modeling have opened new avenues for tackling the scalability challenges of PFAS detection. 
For example, DeLuca et al. \cite{DeLuca_Mullikin_Brumm_2023} used random forests to predict PFAS contamination in fish tissue in the Columbia River Basin, leveraging geospatial data from industrial and military facilities. 
On a national scale, the USGS applied XGBoost to forecast PFAS occurrence in groundwater \cite{Tokranov2023} using PFAS sources, groundwater recharge, etc. 
Such approaches require extensive manual feature engineering, such as aggregating data spatially before prediction is possible (Fig. \ref{fig:test4}). 
In addition to being resource-intensive (see Section 5), this preprocessing loses spatial context. 

\noindent\textbf{Geospatial Models.}
Recent advances in geospatial deep learning have introduced foundation models trained on large-scale satellite datasets for improved spatial prediction. Neural operators and physics-informed models target PDE systems \cite{Li_Kovachki_Azizzadenesheli_Liu_Bhattacharya_Stuart_Anandkumar_2020, Karniadakis_Kevrekidis_Lu_Perdikaris_Wang_Yang_2021}; however, PFAS transport lacks reliable governing physics and dense simulations, motivating a noise-aware, data-driven geospatial approach. Models like Prithvi \cite{Blumenfeld_2023} and SatMAE \cite{Cong_Khanna_Meng_Liu_Rozi_He_Burke_Lobell_Ermon_2023} use masked autoencoders (MAE) to learn generalizable spatial representations in a semi-supervised manner, leveraging both labeled and unlabeled data. Unlike traditional ML approaches that require extensive feature engineering (e.g., Fig. \ref{fig:test4}), these models process raster (image) data directly, preserving spatial dependencies. 
However, many of these approaches assume that  key information is directly observable in satellite imagery, which is not the case for PFAS pollution (see Section 5). We therefore adapt Prithvi to use preprocessed, proxy features, such as land cover. 

\noindent\textbf{Addressing Data Scarcity.}
In environmental data analysis, limited ground truth often necessitates pseudo- or augmented labels, which can introduce noise and lead to overfitting \cite{Safonova_2023}. Various strategies have been proposed to mitigate these issues, including label smoothing and robust loss functions (e.g., Huber loss \cite{Gokcesu_Gokcesu_2021}, Generalized Cross-Entropy \cite{Zhang_Sabuncu_2018}, bootstrapping loss \cite{Reed_Lee_Anguelov_Szegedy_Erhan_Rabinovich_2015}), and focal loss \cite{Lin_Goyal_Girshick_He_Dollár_2018}). Among these, focal loss is particularly effective for addressing class imbalance by down-weighting well-classified examples and focusing the model's learning on hard, uncertain samples.  Weakly supervised methods like the FESTA loss \cite{Hua_Marcos_Mou_Zhu_Tuia_2022} further aim to capture spatial and feature relationships. 
FESTA  is specifically designed for segmentation of remote sensing images using sparse annotations assuming 
spatial continuity, which may not hold for PFAS (see Section 5). 
%
Moreover, limited understanding of PFAS hydrological processes complicates the use of transfer and active learning, making it difficult to estimate uncertainty and identify reliable proxies \cite{Guelfo_2021, Russo_2020}.
%
%
These challenges underscore the need for methods that integrate domain-specific knowledge directly into the training process.
 \begin{figure}[t]
  \centering
  \includegraphics[width=0.9\columnwidth]{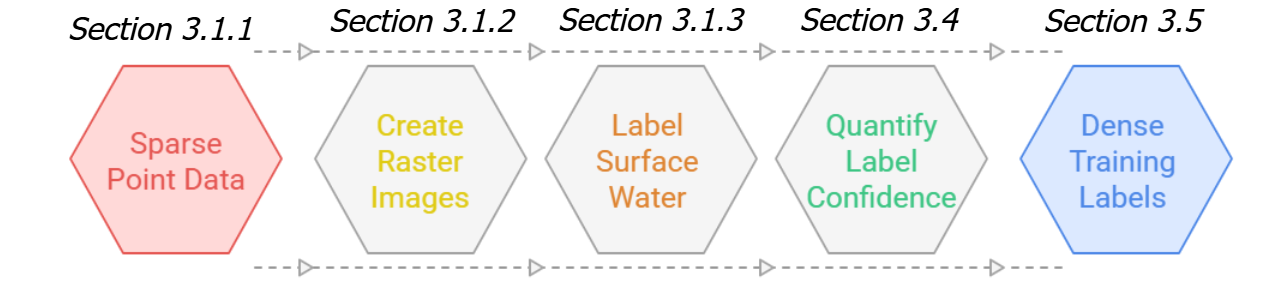}
  \caption{Overview of dataset curation pipeline}
  \label{fig:data_curation_overview}
\end{figure}
\section{Methodology}\label{sec:methodology}


PFAS mapping presents unique challenges: labels exist only at sparse monitoring locations, yet contamination spreads via hydrological and land-use processes. To address this, we propose \pname as a geospatial learning framework with a noise-aware objective grounded in a latent contamination model and hydrology-informed pixel correctness priors. We perform segmentation on surface water area, classifying them as having PFAS concentrations above or below established health advisory thresholds using geospatial data.

\subsection{Dataset}
To contextualize our data preparation pipeline, Fig.~\ref{fig:data_curation_overview} illustrates how ground truth samples are integrated with geospatial data to create the final dataset.

\subsubsection{Ground Truth Samples} We use a combination of datasets and health advisories from government agencies in the United States, for both water and fish tissue samples. These are summarized in Table \ref{tab:dataset_summary} (details in appendix).

\begin{table}[H]
\centering
\caption{Datasets used for training, testing, and real-world validation.}
\label{tab:dataset_summary}
\setlength{\tabcolsep}{5pt}
\renewcommand{\arraystretch}{1.15}
\begin{tabularx}{\columnwidth}{l l X r}
\toprule
\textbf{Type} & \textbf{Use} & \textbf{Data Source} & \textbf{Total Size} \\
\midrule
Fish  & Train/Test  & EPA \cite{USEPA2015,US_EPA_2024} & 866 \\
Fish  & Validation  & MPART \cite{egle_fcmp_2025}      & 114 \\
Water & Train/Test  & MPART \cite{egle_pfas_surface_water} & 293 \\
Water & Validation  & Sampled by Authors              & 8 \\
\bottomrule
\end{tabularx}
\end{table}


We use both fish and water samples: fish reflect longer-term accumulation relevant to exposure risk, while water samples capture localized contamination at a given time \cite{EWGstudy2023}. Our analysis primarily uses fish data due to the nationally representative extent. 


Furthermore, PFAS concentrations are measured as continuous values in these samples, with individual measurements for different species of PFAS. We binarize PFAS concentration measurements into above (1) or below (0) health advisory thresholds, such as the EPA's Fish and Shellfish Advisory Program health thresholds \cite{Contaminants_to_Monitor_in_Fish_and_Shellfish_Advisory_Programs_2024}. In particular, we use the 
cumulative metric called the Hazard Index, which assesses combined risk from individual PFAS compounds by summing their concentration-to-threshold ratios \cite{epa_hazard_index_2023}. Note that for fish train/test data, this yields 775 above-threshold  (89.5\%) and 91 below-threshold (10.5\%) samples, reflecting a critical real-world imbalance. For completeness, we also experimented with an alternative labeling scheme using three classes (see appendix).

\begin{figure*}[t]
  \centering
  \includegraphics[width=0.70\textwidth]{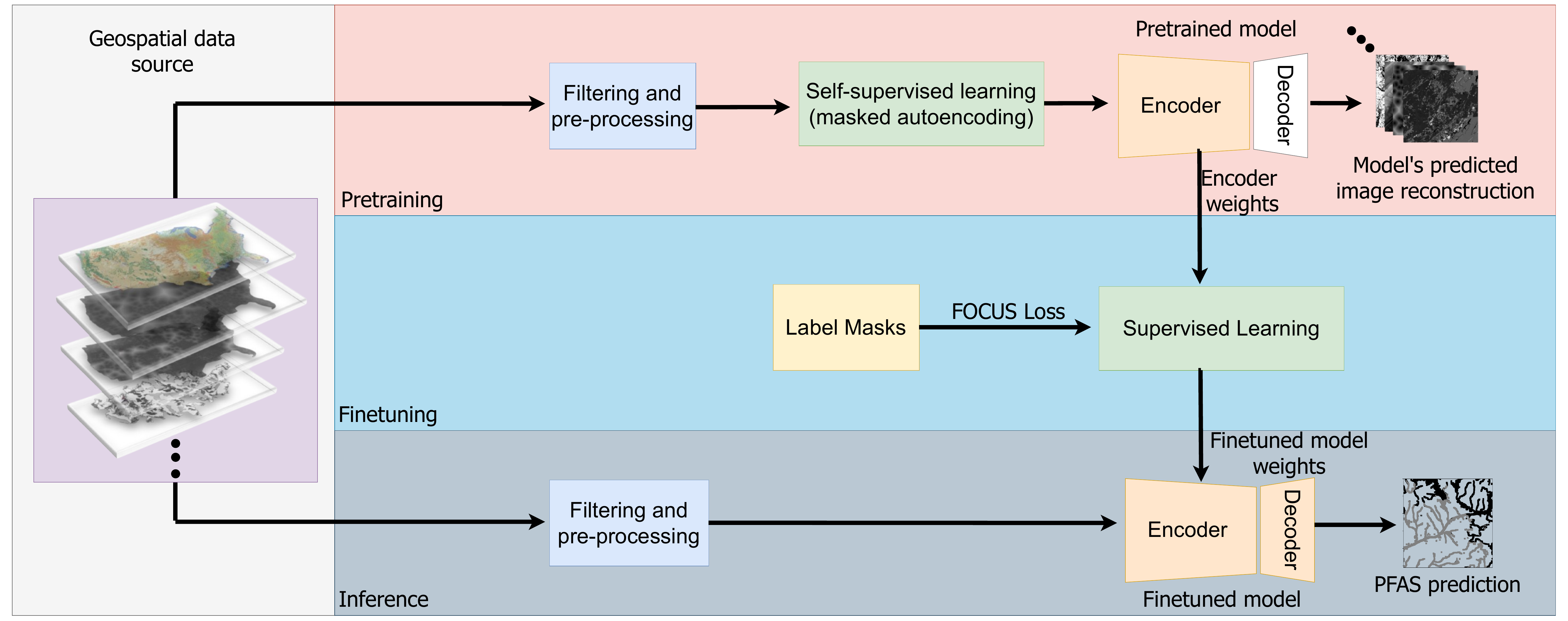}  
  \caption{Model overview: high-level representation of \pname}
  \label{fig:your-image-label2} 
\end{figure*}

\subsubsection{Raster Image Generation}
\begin{figure}[!h]
  \centering
  \includegraphics[width=0.75\columnwidth]{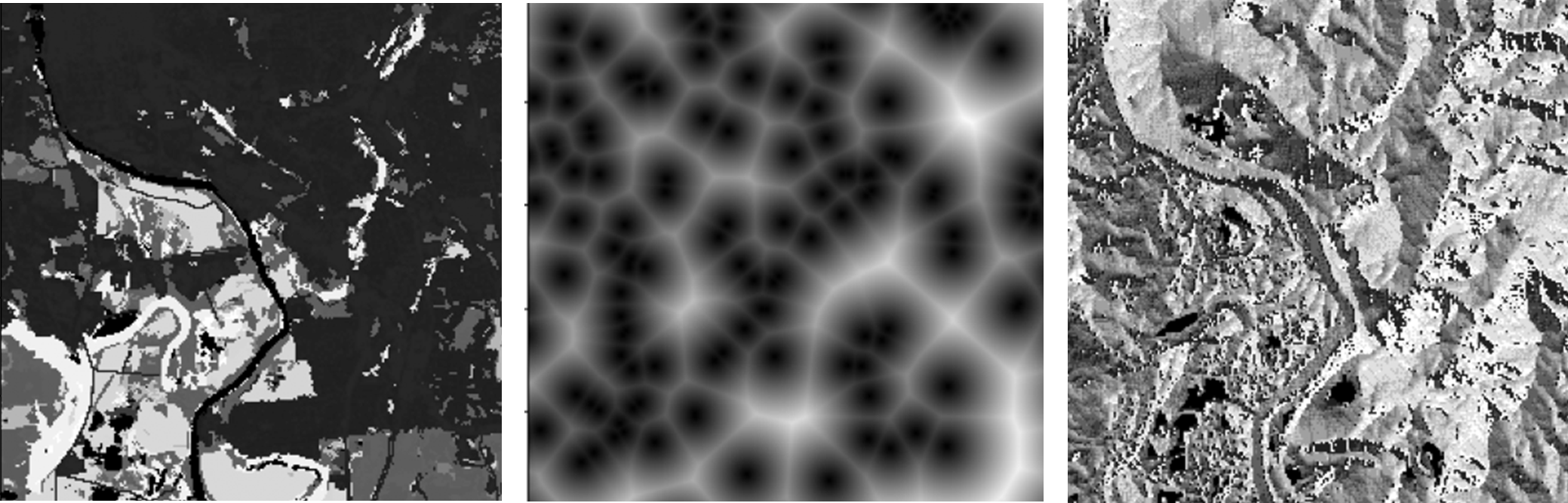}
  \caption{Example raster channels: (left) land cover, (center) distances from  
  chemical manufacturing industries, and (right)  flow direction. }
  \label{fig:chans}
\end{figure}

To spatially integrate the data for training and testing, multi-channel raster images were generated around each sample point (see Fig~\ref{fig:chans}). 
Each image is a \(P \times P\) pixels patch with  30-meter resolution, centered on the geographic coordinates of a sample point, where the optimal value of \(P\) is determined in Section 4.3. Once trained, the model can make predictions at any location simply by extracting and processing the corresponding raster, which is done for real-world validation. 

These raster images integrate several key features. First, we incorporate readily available rasters such as the National Land Cover Database (NLCD)  \cite{USGS_NLCD_2023}, and flow direction rasters that capture hydrological connectivity using the D8 algorithm in ArcGIS Pro \cite{ArcGIS_FlowDirection}. In addition, we 
include discharger location data  obtained from the U.S. EPA Enforcement and Compliance History Online (ECHO) \cite{US_EPA_Water_Pollution} and convert them into distance rasters using a distance transform \cite{scipy_distance_transform_edt}. 

\subsubsection{Ground Truth Masks}
Our goal is to perform segmentation to predict PFAS contamination in surface water, but we have sparse point-level contamination measurements rather than dense labels across rasters. To generate dense training labels, 
every surface water pixel in a patch is labeled according to the point-level sample measurement in that patch (i.e., 
if the sample point indicates PFAS above the safety thresholds, all surface water pixels are labeled 1; if below, they are labeled 0). Non-surface water pixels are assigned a value of 2. This approach extends point-level data across the patch, assuming that nearby areas may have similar contamination.

\subsection{Problem Setup}
Let $z_i \in \{0,1\}$ denote the latent contamination state at pixel $i$, and $y_i \in \{0,1\}$ be the observed label obtained by expanding point measurements across surface water (see Section 3.1.3). Given pixel features $x_i$ (land cover, hydrology, distance to facilities, etc.), the goal is to learn
\begin{equation}
\label{eq:model_prob}
\pi_\theta(x_i)
= \Pr\!\left(z_i = 1 \mid x_i\right)
\end{equation}

under sparse and noisy supervision.

Two challenges arise:

\begin{enumerate}[leftmargin=4mm, topsep=0pt]
    \item \textbf{Spatially-structured label noise:} contamination spreads directionally (downstream), is source-driven (industrial discharge), and varies by land use.
    \item \textbf{Regulatory sensitivity:} false negatives pose public health risk, requiring calibrated uncertainty and high recall on positives.
\end{enumerate}

\subsection{Latent Noise Model}
We assume per-pixel asymmetric flip noise $\eta_i < \frac{1}{2}$, where $\eta_i = \Pr(y_i \neq z_i \mid x_i)$, motivated by laboratory detection limits and geographic label propagation. The noisy likelihood is:

\begin{equation}
\label{eq:noisy_prob}
\Pr\!\left(y_i = 1 \mid x_i\right)
= \eta_i + (1 - 2\eta_i)\,\pi_\theta(x_i)
\end{equation}

Unlike classical noisy-label models that assume global or class-conditional noise rates \cite{Natarajan_Dhillon_Ravikumar_Tewari_2013,Goldberger_Ben-Reuven_2017}, we model 
$\eta_i$ as pixel-wise and spatially varying, reflecting physical measurement limits and label propagation rather than annotation randomness. While instance-dependent noise has been studied (e.g., Confident Learning \cite{Northcutt_Jiang_Chuang_2021}), here $\eta_i$ is driven by geophysical processes and spatial structure, not inferred solely from prediction statistics.


\subsection{Physically-Informed Pixel Confidence $M_i$ Details}

We estimate a correctness score $M_i \in [0,1]$ for each training pixel using environmental priors:

\begin{equation}
\label{eq:Mi_def}
M_i
= \alpha_1\, p_{\text{discharger}}
+ \alpha_2\, p_{\text{landcover}}
+ \alpha_3\, p_{\text{sample}}
+ \alpha_4\, p_{\text{downstream}}
\end{equation}


where:

\begin{itemize}[leftmargin=4mm, topsep=0pt]
    \item $p_{\text{discharger}}$ decays with distance to PFAS facilities,
    \item $p_{\text{landcover}}$ reflects industrial vs. natural land types,
    \item $p_{\text{sample}}$ decays from monitoring points,
    \item $p_{\text{downstream}}$ increases along hydrological flow.
\end{itemize}

These terms map environmental processes to \emph{pixelwise label correctness priors}.  
Weights $(\alpha_1,\alpha_2,\alpha_3,\alpha_4)$ are selected based on domain knowledge and experimental validation. 

We perform an ablation study on the 2008 PFAS dataset using a grid search over 24 noise-mask configurations derived from four inputs: dischargers, land cover, flow direction, and distance to sample points. We vary their relative weights and evaluate segmentation performance across multiple metrics. The best configuration assigns 40\% to dischargers, 20\% to land cover, 10\% to sample distance, and 30\% to flow direction. These results align with domain expert guidance that emphasizes the importance of industrial activity in driving contamination patterns. Pixels at
ground truth sample points labeled 1 are assigned a probability of 1, reflecting complete confidence. However, for sample points labeled 0, we account for additional uncertainty due to certain PFAS compounds having method detection limits (MDLs) that exceed advisory thresholds\footnote{In such cases, a compound may go undetected despite being present at a high-risk level. To address this, the corresponding noise mask value is down-weighted in proportion to the ratio of the advisory threshold to the MDL}.
Full performance results for all configurations and additional details on the environmental priors are reported in the appendix. 

\subsection{\pname Loss}
We scale focal loss by $M_i$ to emphasize trustworthy pixels and hard cases:

\begin{equation}
\label{eq:focus_loss}
\mathcal{L}_{\text{FOCUS}}
= \frac{1}{N} \sum_{i=1}^{N}
M_i (1 - p_i)^\gamma
\Big[
- y_i \log p_i
- (1 - y_i)\log(1 - p_i)
\Big]
\end{equation}

where \( N \) is the total number of pixels, and \( p_i \) is the predicted probability of the positive class. 
Key components:

\begin{itemize}[leftmargin=4mm, topsep=0pt]
\item $M_i$ down-weights uncertain pixels (noise-robust)
\item $(1-p_i)^\gamma$ up-weights hard examples (focal modulation)
\end{itemize}

\begin{tcolorbox}[
  title={Proposition 1 (Noise-Aware Surrogate Objective)},
  colback=gray!5,
  colframe=black,
  boxrule=0.5pt,
  sharp corners
]
Let $z_i \in \{0,1\}$ denote the latent contamination state and
$y_i \in \{0,1\}$ the observed expanded label under a pixel-specific
asymmetric flip noise rate $\eta_i < \tfrac{1}{2}$, such that
\[
\Pr(y_i = 1 \mid x_i) = \eta_i + (1 - 2\eta_i)\,\pi_\theta(x_i).
\]
Then there exist coefficients $a_i, b_i$ such that the noisy
negative log-likelihood satisfies
\[
\ell_{\text{noisy}}(p_i, y_i) \le a_i \cdot \mathrm{CE}(p_i, y_i) + b_i.
\]
where $\mathrm{CE}$ denotes the standard cross-entropy loss. Setting $a_i = M_i$, where $M_i$ increases as $\eta_i$ decreases,
yields a valid surrogate objective whose expectation upper-bounds
the $\ell_{\text{noisy}}(p_i, y_i)$ and serves as a lower-bound surrogate to the clean
log-likelihood.
\end{tcolorbox}

\noindent\textbf{Proof sketch.}
Under the asymmetric flip noise model with pixel-specific noise rate
$\eta_i < \tfrac{1}{2}$, the noisy likelihood satisfies
\[
\Pr(y_i = 1 \mid x_i) = \eta_i + (1 - 2\eta_i)\,p_i,
\]
where $p_i = \pi_\theta(x_i)$. The corresponding negative log-likelihood is
\[
\ell_{\text{noisy}}(p_i, y_i) = -\log\!\left(\eta_i + (1 - 2\eta_i)\,p_i\right).
\]
Since the logarithm is concave, for any fixed anchor $p_0 \in (0,1)$
there exist coefficients $a_i, b_i$ such that
\[
\ell_{\text{noisy}}(p_i, y_i) \le a_i \cdot \mathrm{CE}(p_i, y_i) + b_i,
\]
We then apply the standard focal modulation $(1-p_i)^\gamma$, where $\gamma \ge 0$ controls the focus on hard examples, to the CE term.




If we set $a_i = M_i$, where $M_i$ increases as $\eta_i$ decreases (i.e., higher label trust), then $\mathcal{L}_{\text{FOCUS}}$
is a valid upper-bounding surrogate of $\mathbb{E}[\ell_{\text{noisy}}]$ (up to additive constants).

Consequently, minimizing $\mathcal{L}_{\text{FOCUS}}$ maximizes a lower bound on the (noisy) log-likelihood, and—under the alignment assumption on $M_i$—serves as a principled surrogate for learning from the underlying clean labels while encoding physically grounded priors rather than learned heuristics. Further details and a formal derivation are provided in the appendix.

\subsection{Model}

Our framework, \pname, is inspired by the Prithvi architecture \cite{Blumenfeld_2023} and employs a masked autoencoder (MAE) approach for pretraining (Fig. \ref{fig:your-image-label2}). Instead of relying on Prithvi’s pretrained weights, we pretrain our model on derived geospatial data products (e.g., land cover and distance rasters), which we hypothesize offer more contextual information for capturing the environmental nuances critical to PFAS contamination prediction (see Section 5). 

\section{Experiments}\label{sec:baselines}
\subsection{Baselines}
To contextualize our proposed framework for PFAS contamination prediction, we introduce several baseline approaches commonly employed in the field:

\begin{itemize}
    \item \textbf{Pollutant transport simulation:} A scientific, process-based approach that models contaminant movement using hydrological principles \cite{SWAT2023}. Further details on implementation in appendix.
    \item \textbf{Landsat-based method:} Uses the original Prithvi weights to predict PFAS from raw satellite imagery (Landsat 7 multispectral  \cite{USGSLandsat7}).
    \item \textbf{FESTA loss approach:} A state-of-the-art method for sparse segmentation from point supervision. 
    \cite{Hua_Marcos_Mou_Zhu_Tuia_2022}.
    \item \textbf{Kriging:} A geostatistical interpolation method \cite{esri_kriging} that predicts contamination based on spatial relationships between points. Further details on implementation in appendix.
    \item \textbf{Random forest:} Closely following the methodology of \cite{DeLuca_Mullikin_Brumm_2023}, this model aggregates environmental features within a 5 km buffer around each sample (e.g., calculating the percentage of various land cover types) 
    to predict contamination at individual points.
    \item \textbf{Rule-based approach:} 
    Assigns contamination labels based on predefined environmental heuristics only, replicating our noise mask approach to compute a contamination probability (but without (\(p_{\text{sample\_dist}}\))), which is then thresholded to yield binary labels. 
     
\end{itemize}


\subsection{Training Configuration}


We now describe our experimental setup, with additional details in the appendix.
We conduct experiments \textit{per year} to evaluate \textbf{the model’s ability to fill spatial data gaps, an important stakeholder need.} We do not forecast across time as longitudinal ground truth is currently too sparse and inconsistent for reliable temporal supervision. For each year, we split the data into 80\% training and 20\% testing, with geographically disjoint regions to avoid spatial leakage and ensure diversity.

We use a batch size of 4 to preserve the stability of the pretrained backbone and optimize with AdamW \cite{Loshchilov_Hutter_2019} (
\(\beta_1 = 0.9\) and \(\beta_2 = 0.999\)). A custom learning-rate schedule with warmup followed by polynomial decay stabilizes convergence and enables gradual adaptation to limited data. Hyperparameters were selected based on consistent convergence.

\begin{figure*}[t]
  \centering
  \includegraphics[width=0.9\textwidth]{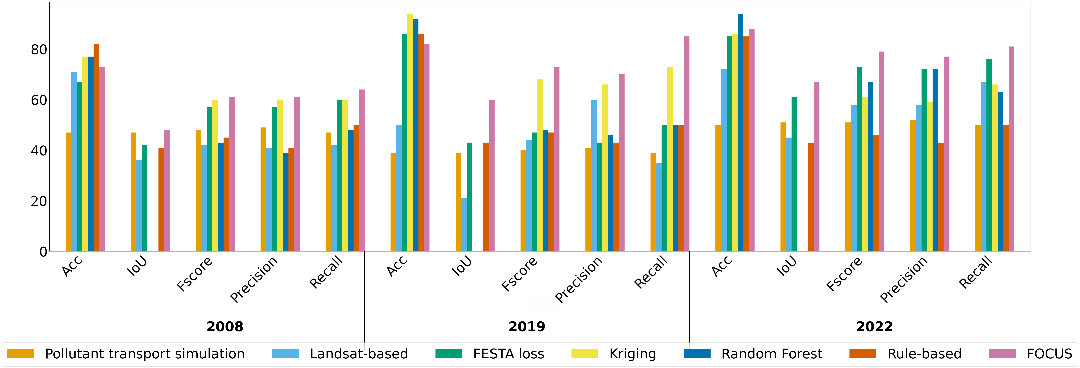}  
  \caption{Comparison of performance across different methods and years. Results are averaged over 3 random seeds for \pname and applicable baselines (Landsat-based, FESTA loss, rule-based). \emph{IoU is not applicable to methods that do not produce dense pixel-level segmentations (e.g., Kriging and Random Forest), and is therefore omitted for those approaches.}}
  \label{fig:your-image-label}
\end{figure*}

\subsection{Ablation Study}


We conduct ablation studies across the following dimensions:

\begin{itemize}
   
    \item \textbf{Noise Weights:} We evaluate the impact of incorporating noise masks into the \pname loss function. With noise masks, pixel-wise confidence scores scale focal loss to emphasize high-confidence pixels; without them, the loss reduces to standard focal loss  \cite{Lin_Goyal_Girshick_He_Dollár_2018}.

     \item \textbf{Raster Size:} To understand the influence of spatial context, we compare raster patches of \(512 \times 512\) and \(256 \times 256\) pixels. Larger patches capture broader context, potentially including contamination sources farther from the sample point, while smaller patches reduce label noise and focus on localized features. 
\end{itemize} 

While the primary focus of this work is on PFAS contamination, we also explored the generalizability of our framework to land cover classification, and robustness through a consistency analysis detailed in the appendix.


\subsection{Metrics}

The reported metrics include accuracy, Intersection over Union (IoU), F-score, precision, and recall. Performance metrics in this study are reported exclusively from evaluations \textbf{using only ground truth samples}. 
In addition, we assess timing efficiency (in seconds) as a metric to evaluate the model's computational scalability. 

For threshold-dependent metrics, we select the decision threshold using out-of-fold predictions on the training data. Specifically, we perform $K$-fold cross-validation on the training set, sweep thresholds to maximize F-score on the resulting out-of-fold predictions,
and apply the selected threshold to the held-out test set. We use $K=5$ folds and sweep 200 evenly spaced thresholds in $[0,1]$.
Thresholds are selected independently for each method.


\section{Results}\label{sec:results}


\subsection{\pname Outperforms Existing Baselines}

\subsubsection{Main Results}
We first compared \pname against six alternative approaches for PFAS contamination prediction, as shown in Fig.~\ref{fig:your-image-label}. \textbf{Overall, \pname provides the most consistent performance across years and metrics.}. While several baselines achieve competitive accuracy, \textbf{according to domain experts, accuracy alone is insufficient in this highly imbalanced setting, making precision–recall trade-offs more informative}. 
In particular, Kriging and the FESTA loss achieve comparable accuracy in some years and show reasonable performance across metrics, highlighting the strength of spatial interpolation and structure-aware learning in this setting. The Landsat-based model yields moderate results, suggesting that satellite features alone are insufficient without explicit geospatial context. The random forest approach, despite achieving acceptable accuracy, suffered from  lower F-score, precision, and recall, likely due to the limited spatial context from per-point feature aggregation. Pollutant transport simulation and the rule-based approach provide useful physics- and heuristic-driven references (the latter also serving as a component in our noise modeling), but they lag behind data-driven methods in overall predictive balance. Taken together, these results indicate that \pname consistently achieves a favorable trade-off between precision and recall across years, which is critical for robust large-scale environmental screening.

\subsubsection{Time Efficiency}

We compare \pname against
a random forest baseline (following DeLuca et al.~\cite{DeLuca_Mullikin_Brumm_2023}) using similar environmental features and datasets. Table~\ref{tab:performance_comparison_ml_dl} reports feature extraction and inference times over a 1 km$^2$ area and all of Northern Michigan (NM, $\sim$44{,}000 km$^2$).
The ML baseline requires aggregating features within a spatial buffer per point, resulting in significantly higher extraction costs at scale. In contrast, \pname processes full raster patches efficiently, reducing feature extraction time from days to hours over NM, while inference times remain comparable between methods. Feature extraction was performed on an Intel Xeon Gold 6154 CPU (3.0 GHz), and inference used an AMD Ryzen Threadripper PRO 7985WX with an NVIDIA RTX 6000 Ada GPU.
\begin{table}[ht!]
  \centering
  \caption{Comparison of performance for feature extraction and inference using random forest and \pname over 1 km\(^2\) area and Northern Michigan (NM). Results averaged over 10 runs.}
  \resizebox{\columnwidth}{!}{  
    \setlength{\tabcolsep}{5pt}  
    \renewcommand{\arraystretch}{1.3}  
    \begin{tabular}{l c c c c c c}
      \toprule
      & \multicolumn{2}{c}{\textbf{Feature extraction}} & \multicolumn{2}{c}{\textbf{Inference}} \\
      \cmidrule(lr){2-3}\cmidrule(lr){4-5}
      & \textbf{ML} & \textbf{DL} & \textbf{ML} & \textbf{DL} \\
      \midrule
      For 1 km\(^2\) area & 1.2 $\pm$ 0.2 mins & 4.7 $\pm$ 0.5 sec & 0.0003 $\pm$ 0.0002 sec & 0.0005 $\pm$ 0.0001 sec \\
      For NM & 2 days & 3.2 hrs $\pm$ 0.4 hrs & 7.5 $\pm$ 1 sec & 13 $\pm$ 2 sec \\
      \bottomrule
    \end{tabular}
  }
  \captionsetup{justification=centering, position=bottom}  
  \label{tab:performance_comparison_ml_dl}
\end{table}

\subsection{\pname Applies in the Real World}



\begin{figure}[ht!]
  \centering
  \includegraphics[width=0.75\columnwidth]{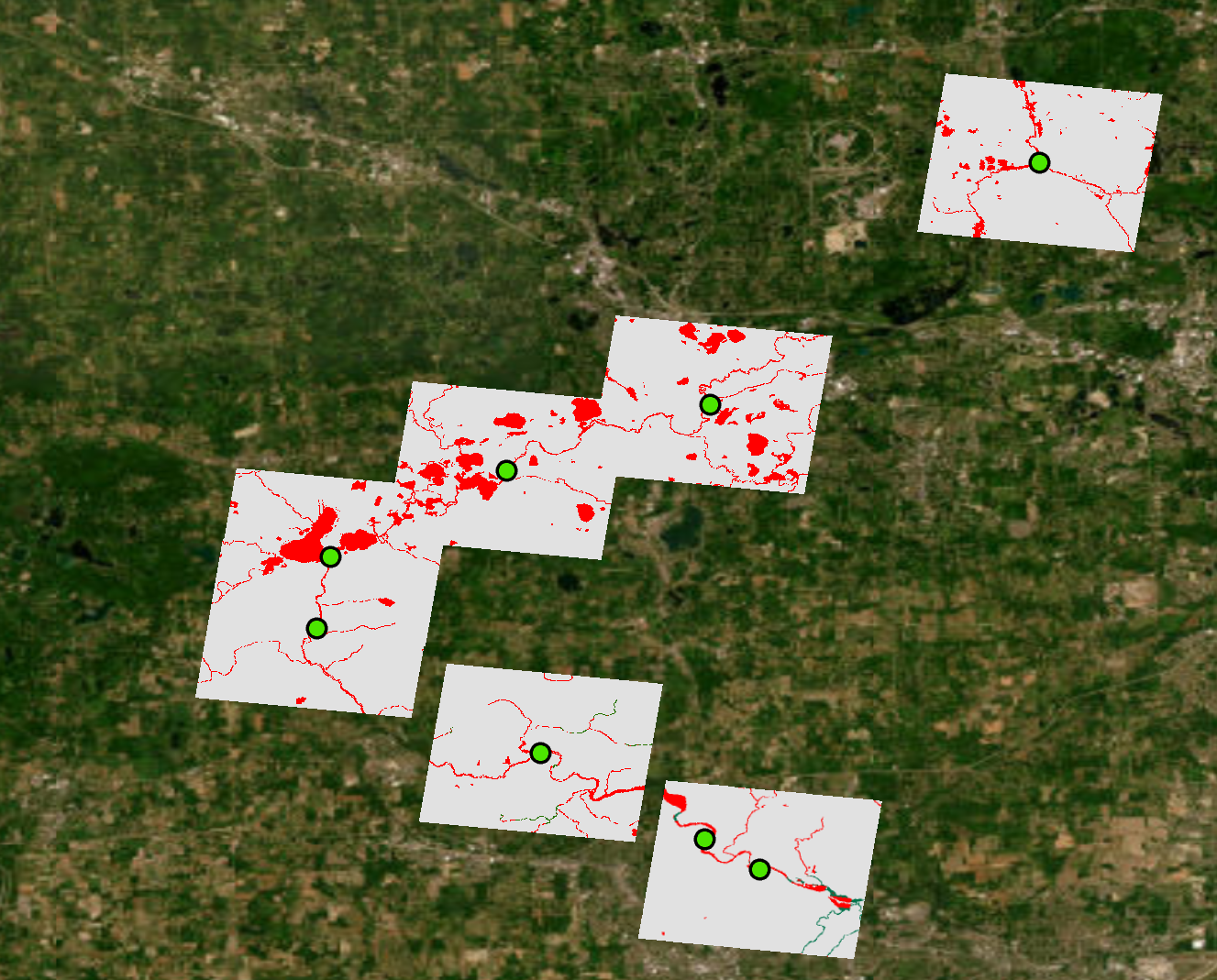}
  \caption{Real-world validation using surface water samples collected by our team. Newly sampled locations (green) are overlaid on model predictions, with red indicating areas of high predicted contamination, and dark green indicating areas of lower predicted contamination.}
  \label{fig:test2}
\end{figure}

We next conducted two complementary validation efforts of predicted contamination maps \textbf{of new regions} -- beyond regions surrounding ground truth sample points in the train and test sets -- using field-collected data (see Table \ref{tab:dataset_summary}). First, our research team collected surface water samples from multiple local sites in Ann Arbor, Michigan, in mid-2025. These sites represent previously unsampled regions and provide an independent assessment of model generalization beyond our test set under sparse and biased sampling, as testing is typically limited due to cost and feasibility constraints. Notably, as shown in Fig.~\ref{fig:test2}, the model, trained on 2024 surface water data from MPART, correctly identified the 8 collected points as highly contaminated, demonstrating FOCUS’s ability to generalize under severe label scarcity. We are actively pursuing larger-scale validation with stakeholders and regulators. Technical details on sample analysis are provided in the appendix. 
Second, we evaluated our 2019 fish tissue model in Table \ref{tab:realworld_val} on the publicly available MPART fish dataset from 2019 \cite{egle_fcmp_2025}, which includes PFAS measurements in fish tissue from across Michigan, collected independently from our fish train and test sets. Results are comparable to Fig. \ref{fig:your-image-label}. 

\begin{figure}[!t]
\centering
\begin{subfigure}[t]{0.30\columnwidth}
  \centering
  \fbox{\includegraphics[width=\linewidth]{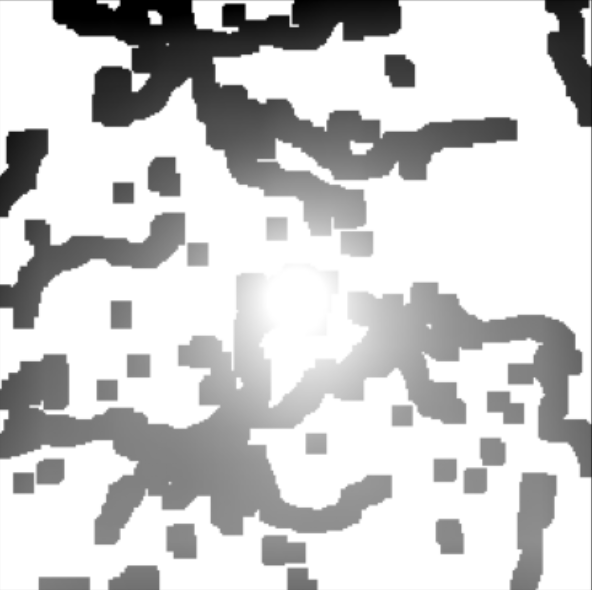}}
  \caption{Pixelwise confidence $M_i$ (white = higher trust)}
\end{subfigure}
\hfill
\begin{subfigure}[t]{0.30\columnwidth}
  \centering
  \fbox{\includegraphics[width=\linewidth]{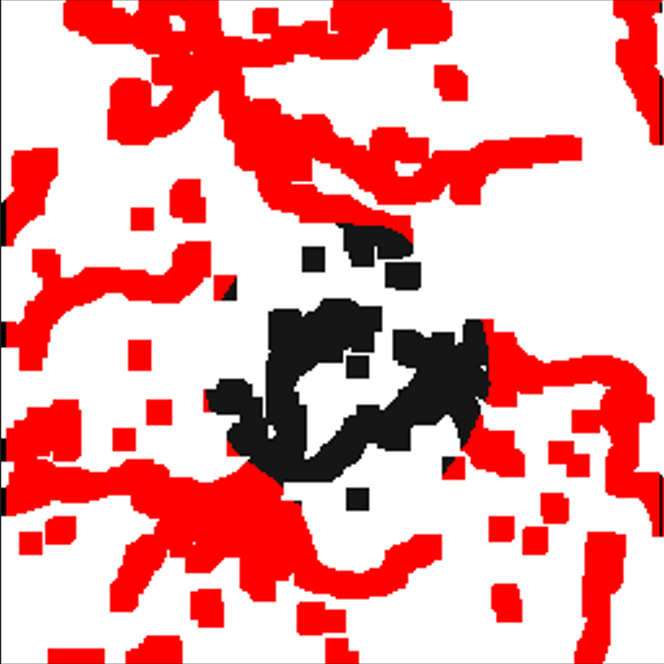}}
  \caption{Focal probability map}
\end{subfigure}
\hfill
\begin{subfigure}[t]{0.30\columnwidth}
  \centering
  \fbox{\includegraphics[width=\linewidth]{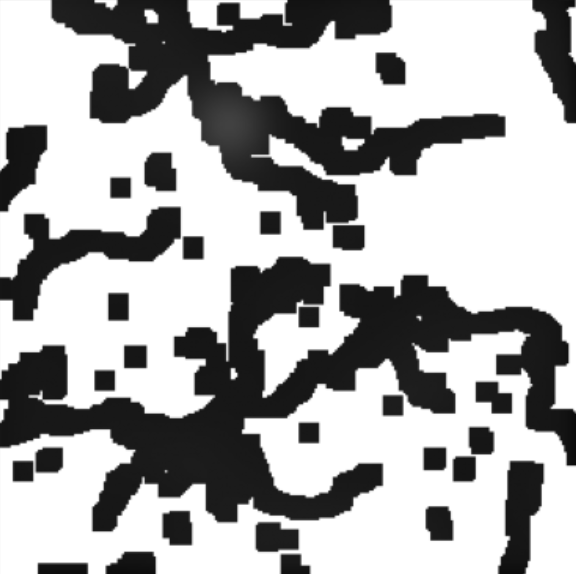}}
  \caption{\pname probability map}
\end{subfigure}

\caption{Grayscale PFAS probabilities (white = higher). Red overlays indicate overconfident positives in low-trust regions $(p_i \ge 0.50,\; M_i \le 0.30)$. \pname suppresses such responses, while the focal baseline produces many.}
\label{fig:error_maps}
\end{figure}

Qualitatively, Fig.~\ref{fig:error_maps} shows grayscale probability maps with red overlays indicating confident positive predictions in low-trust regions; the focal baseline produces many such responses, whereas \pname largely suppresses them, yielding more conservative behavior under unreliable supervision.


\begin{table}[h]
  \centering
  \caption{\pname performance on fish tissue samples collected independently of training data. Results  averaged over 3 random seeds.}
  \resizebox{1.0\columnwidth}{!}{
    \setlength{\tabcolsep}{6pt}
    \renewcommand{\arraystretch}{1.2}
    \begin{tabular}{lccccc}
      \toprule
      \textbf{Metric} & \textbf{Accuracy} & \textbf{IoU} & \textbf{F1} & \textbf{Precision} & \textbf{Recall} \\
      \midrule
      MPART Fish Samples & 85 $\pm$ 2 \% & 60 $\pm$ 1 \% & 72 $\pm$ 1 \% & 75 $\pm$ 1 \% & 70 $\pm$ 2 \% \\
      \bottomrule
    \end{tabular}
  }
  \label{tab:realworld_val}
\end{table}


\subsection{\pname Noise Masks Improve Performance}\label{sec:noise}

\begin{table}[ht!]
  \centering
  \caption{Ablation showing impact of noise-aware loss (\pname) vs. standard focal loss. Results  averaged over 3 random seeds. Bold values denote the best result per year.}
  \resizebox{\columnwidth}{!}{
    \setlength{\tabcolsep}{5pt}
    \renewcommand{\arraystretch}{1.3}
    \begin{tabular}{l c c c c c c}
      \toprule
      & \multicolumn{2}{c}{\textbf{2008 (\%)}} & \multicolumn{2}{c}{\textbf{2019 (\%)}} & \multicolumn{2}{c}{\textbf{2022 (\%)}} \\
      \cmidrule(lr){2-3} \cmidrule(lr){4-5} \cmidrule(lr){6-7}
      \textbf{Metric} & \textbf{Focal only} & \textbf{\pname} & \textbf{Focal only} & \textbf{\pname} & \textbf{Focal only} & \textbf{\pname} \\
      \midrule
      Accuracy  & 37 $\pm$ 2 & \textbf{73 $\pm$ 1} & 62 $\pm$ 1 & \textbf{82 $\pm$ 1} & 55 $\pm$ 3 & \textbf{88 $\pm$ 2} \\
      IoU       & 22 $\pm$ 3 & \textbf{48 $\pm$ 2} & 41 $\pm$ 2 & \textbf{60 $\pm$ 2} & 36 $\pm$ 3 & \textbf{67 $\pm$ 2} \\
      F-score   & 36 $\pm$ 3 & \textbf{61 $\pm$ 2} & 57 $\pm$ 1 & \textbf{73 $\pm$ 2} & 53 $\pm$ 3 & \textbf{79 $\pm$ 2} \\
      Precision & 54 $\pm$ 2 & \textbf{61 $\pm$ 2} & 63 $\pm$ 2 & \textbf{70 $\pm$ 2} & 63 $\pm$ 3 & \textbf{77 $\pm$ 2} \\
      Recall    & 55 $\pm$ 2 & \textbf{64 $\pm$ 2} & 78 $\pm$ 1 & \textbf{85 $\pm$ 1} & 74 $\pm$ 1 & \textbf{81 $\pm$ 2} \\
      \bottomrule
    \end{tabular}
  }
  \label{tab:ablation_noise}
\end{table}

Table~\ref{tab:ablation_noise} evaluates the effect of incorporating noise masks in \pname, compared to focal loss alone (i.e., removing $M_i$ from Eq.~4). Adding noise masks consistently improves performance by emphasizing high-confidence pixels and down-weighting uncertain ones during training. This ablation uses a \(256 \times 256\) resolution, selected based on the patch-size study in the appendix, where it outperformed \(512 \times 512\), likely due to reduced label-noise propagation over large contexts. We did not evaluate smaller patches, as overly small contexts can degrade accuracy and increase variability \cite{Lechner_Stein_Jones_Ferwerda_2009}.

\subsection{\pname has Robust Results}
To evaluate generalization to unseen regions, we perform spatial cross-validation by holding out sets of U.S. states (Table~\ref{tab:cross_state_val}). Specifically, three disjoint groups of five states each are used as test sets, with the remaining states used for training, ensuring no spatial overlap and mimicking deployment in unsampled areas. \pname consistently outperforms focal loss across all years, demonstrating stronger spatial generalization and robustness under distribution shifts.

We also assess spatial robustness via a \textbf{consistency analysis} across all 49 continental U.S. states by comparing predictions on overlapping patches at two overlap scales (56$\times$56 and 156$\times$156 pixels). We observe high agreement (above 93\% on average), indicating locally consistent predictions; full details are provided in the appendix. Results on model calibration (i.e., Expected Calibration Error (ECE) \cite{Guo_Pleiss_Sun_Weinberger_2017}) are also provided in the appendix, showing low ECE values (\(\leq 0.1\)) and indicating that predicted probabilities are well-aligned with observed outcomes.
\begin{table}[!ht]
  \centering
  \caption{Cross-state generalization results (\(\text{mean} \pm \text{std}, \) over the disjoint state splits). Bold values denote the best result per year.}
  \resizebox{\columnwidth}{!}{  
    \setlength{\tabcolsep}{5pt}
    \renewcommand{\arraystretch}{1.3}
    \begin{tabular}{l c c c c c c}
      \toprule
      & \multicolumn{2}{c}{\textbf{2008 (\%)}} & \multicolumn{2}{c}{\textbf{2019 (\%)}} & \multicolumn{2}{c}{\textbf{2022 (\%)}} \\
      \cmidrule(lr){2-3} \cmidrule(lr){4-5} \cmidrule(lr){6-7}
      \textbf{Metric} & \textbf{Focal Only} & \textbf{\pname} & \textbf{Focal Only} & \textbf{\pname} & \textbf{Focal Only} & \textbf{\pname} \\
      \midrule
      Accuracy  & 58 $\pm$ 2 & \textbf{70 $\pm$ 1} & 76 $\pm$ 1 & \textbf{89 $\pm$ 1} & 78 $\pm$ 2 & \textbf{92 $\pm$ 1} \\
      IoU       & 34 $\pm$ 2 & \textbf{48 $\pm$ 1} & 40 $\pm$ 2 & \textbf{55 $\pm$ 1} & 56 $\pm$ 2 & \textbf{75 $\pm$ 1} \\
      F-score   & 47 $\pm$ 1 & \textbf{63 $\pm$ 1} & 46 $\pm$ 2 & \textbf{61 $\pm$ 1} & 68 $\pm$ 1 & \textbf{85 $\pm$ 1} \\
      Precision & 49 $\pm$ 2 & \textbf{62 $\pm$ 1} & 49 $\pm$ 3 & \textbf{59 $\pm$ 2} & 69 $\pm$ 2 & \textbf{82 $\pm$ 1} \\
      Recall    & 49 $\pm$ 2 & \textbf{72 $\pm$ 1} & 50 $\pm$ 1 & \textbf{71 $\pm$ 2} & 85 $\pm$ 1 & \textbf{91 $\pm$ 1} \\
      \bottomrule
    \end{tabular}
  }
  \captionsetup{justification=centering, position=bottom}
  \label{tab:cross_state_val}
\end{table}

\section{Scientific Impact and Interdisciplinary Considerations}

\textbf{This work advances PFAS science by linking large-scale environmental context to surface-water contamination risk and its implications for bioaccumulation.} By integrating industrial dischargers, land cover, and hydrological transport with sparse measurements, our framework produces spatially consistent risk maps that help identify regions likely to be under-sampled, supporting targeted follow-up testing and more systematic risk assessment. At a broader scale, such modeling enables more consistent, nationwide perspectives on contamination risk, complementing existing, state-level fish advisory programs.

\textbf{Beyond static risk mapping, our interface supports scientific exploration and hypothesis generation.} From a scientific perspective, while \pname does not yet perform source attribution, the interface enables exploratory analysis by overlaying predicted risk with known discharger locations and hydrological context. Regions exhibiting high predicted risk despite few nearby known sources, such as the highlighted example in Fig~\ref{fig:webmap}, have also been noted by domain experts as areas of unexplained contamination, suggesting potential unknown, legacy, or poorly characterized sources and motivating targeted follow-up investigation.

\textbf{The need for AI/ML in this domain is driven by severe data sparsity and fragmentation.} PFAS comprise thousands of compounds with diverse sources and uses, yet environmental measurements—especially in surface waters and fish—remain extremely limited in space and time. Traditional monitoring and mechanistic modeling alone cannot provide comprehensive coverage, motivating learning-based approaches that integrate heterogeneous geospatial data and extrapolate risk patterns under uncertainty.

\textbf{Applying AI in this setting also raises scientific and societal challenges, including interpretability, uncertainty communication, and regulatory alignment.} Our approach addresses these concerns by explicitly modeling spatially structured label noise and incorporating physically informed priors, providing more calibrated and transparent predictions than standard black-box models.

\section{\pname Web Map Interface}

\begin{figure}[htbp]
\centering
\includegraphics[width=1.0\columnwidth]{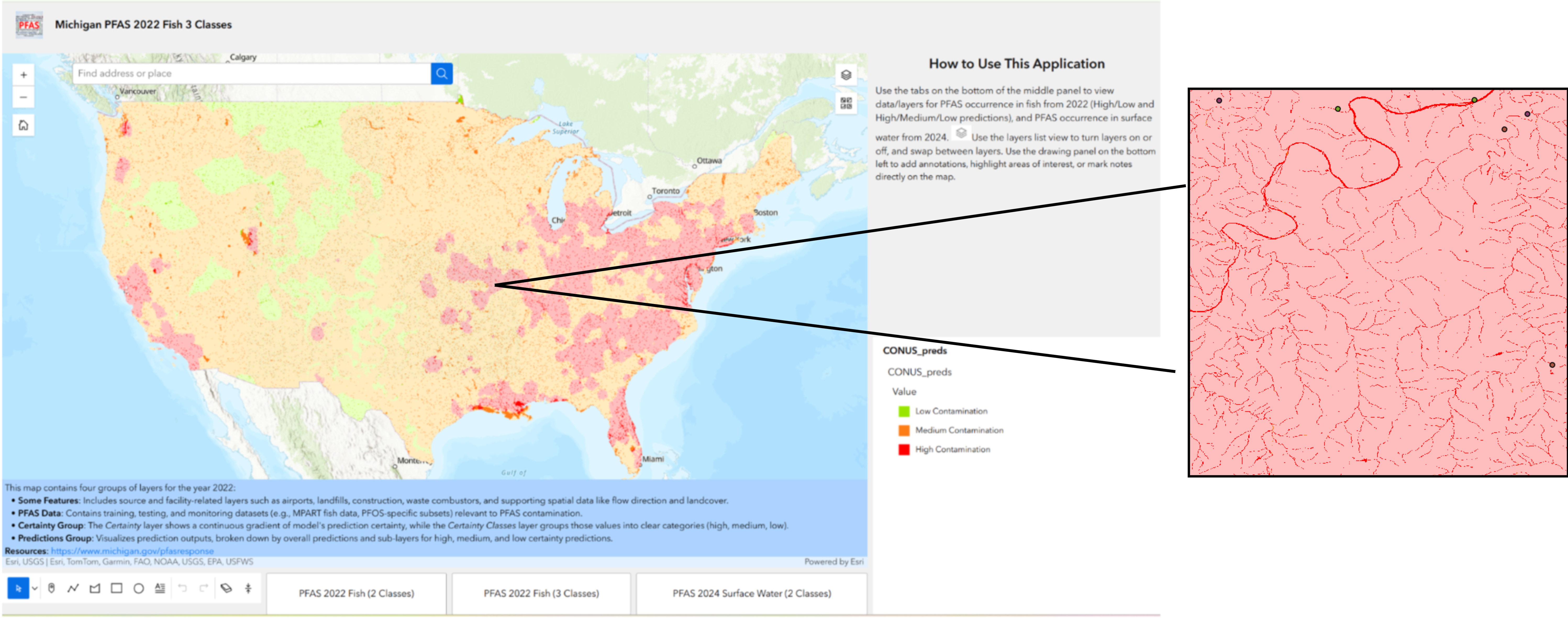}
\caption{Web interface for exploring FOCUS predictions. The zoomed region shows high predicted risk despite few nearby known dischargers (shown as point overlays).}

\label{fig:webmap}
\end{figure}

To support real-world use and stakeholder engagement, we have developed a web-based, interactive PFAS risk mapping interface with policymakers and domain stakeholders that visualizes model predictions at national scale (Fig. \ref{fig:webmap}). The interface displays predicted PFAS contamination risk across surface waters and fish tissue using discrete risk categories (e.g., low, medium, high), enabling intuitive exploration by non-technical users.

The map integrates \pname predictions with contextual geospatial layers, including surface water extents, land cover, hydrological features, and known discharge locations. Users can toggle layers, zoom to regions of interest, and inspect spatial patterns across years and data modalities. The interface was developed in collaboration with policymakers and domain stakeholders to ensure interpretability, transparency, and responsible communication of model outputs.

This deployment illustrates how noise-aware geospatial predictions can be translated into actionable decision-support tools, while acknowledging uncertainty and the limitations of sparse environmental sampling.

\section{Conclusion and Future Work}
We introduced \pname, a geospatial deep learning framework that integrates multi-channel raster inputs with \textbf{a principled noise-aware learning objective} designed for spatially structured label noise arising from point-to-pixel supervision. By combining hydrology-informed correctness priors with a noise-robust loss, \pname improves predictive performance while preserving spatial coherence, enabling large-scale PFAS contamination screening.

Future work will explore predictive uncertainty quantification to guide targeted sampling and enhance prediction robustness. We also aim to incorporate temporal modeling using multi-year data to track and forecast evolving PFAS patterns, as longitudinal measurements become denser and domain understanding enables more physics-informed supervision and priors. Finally, we plan to build on public contamination maps (e.g., EWG \cite{EWG}, \cite{PFAS_tap}) by deploying our framework responsibly. 



\section{Limitations and Ethical Considerations}
As with any large-scale environmental modeling effort, \pname performance is ultimately bounded by the availability and quality of underlying data. Additionally, we consider a static, one-shot prediction setting; coupling \pname with active or adaptive sampling is left to future work. Furthermore, the current model does not yet attribute predicted risk to specific PFAS compounds and sources, though we are actively developing extensions in this direction. Accordingly, predictions should be interpreted as screening-level risk signals to prioritize follow-up sampling rather than definitive contamination assessments.

Given the ethical implications of releasing public-facing PFAS risk maps—including impacts on public health, environmental justice, and property values—we engage an advisory group with representatives from nonprofits and agencies (e.g., HRWC). This collaboration helps tailor outputs to community needs, communicate uncertainty, and support responsible use in remediation and policy decisions.

\bibliographystyle{ACM-Reference-Format}
\bibliography{sample-base}


\begin{thebibliography}{64}


\ifx \showCODEN    \undefined \def \showCODEN     #1{\unskip}     \fi
\ifx \showISBNx    \undefined \def \showISBNx     #1{\unskip}     \fi
\ifx \showISBNxiii \undefined \def \showISBNxiii  #1{\unskip}     \fi
\ifx \showISSN     \undefined \def \showISSN      #1{\unskip}     \fi
\ifx \showLCCN     \undefined \def \showLCCN      #1{\unskip}     \fi
\ifx \shownote     \undefined \def \shownote      #1{#1}          \fi
\ifx \showarticletitle \undefined \def \showarticletitle #1{#1}   \fi
\ifx \showURL      \undefined \def \showURL       {\relax}        \fi
\providecommand\bibfield[2]{#2}
\providecommand\bibinfo[2]{#2}
\providecommand\natexlab[1]{#1}
\providecommand\showeprint[2][]{arXiv:#2}

\bibitem[Blumenfeld(2023)]%
        {Blumenfeld_2023}
\bibfield{author}{\bibinfo{person}{Josh Blumenfeld}.} \bibinfo{year}{2023}\natexlab{}.
\newblock \bibinfo{title}{NASA and IBM Openly Release Geospatial AI Foundation Model for NASA Earth Observation Data | Earthdata}.
\newblock
\urldef\tempurl%
\url{https://www.earthdata.nasa.gov/news/impact-ibm-hls-foundation-model}
\showURL{%
\tempurl}


\bibitem[Bondi-Kelly et~al\mbox{.}(2023)]%
        {Bondi-Kelly_Chen_Golden_Behari_Tambe_2023}
\bibfield{author}{\bibinfo{person}{Elizabeth Bondi-Kelly}, \bibinfo{person}{Haipeng Chen}, \bibinfo{person}{Christopher~D. Golden}, \bibinfo{person}{Nikhil Behari}, {and} \bibinfo{person}{Milind Tambe}.} \bibinfo{year}{2023}\natexlab{}.
\newblock \showarticletitle{Predicting micronutrient deficiency with publicly available satellite data}.
\newblock \bibinfo{journal}{\emph{AI Magazine}} \bibinfo{volume}{44}, \bibinfo{number}{1} (\bibinfo{year}{2023}), \bibinfo{pages}{30–40}.
\newblock
\showISSN{2371-9621}
\href{https://doi.org/10.1002/aaai.12080}{doi:\nolinkurl{10.1002/aaai.12080}}


\bibitem[Breiman(2001)]%
        {Breiman_2001}
\bibfield{author}{\bibinfo{person}{Leo Breiman}.} \bibinfo{year}{2001}\natexlab{}.
\newblock \showarticletitle{Random Forests}.
\newblock \bibinfo{journal}{\emph{Machine Learning}} \bibinfo{volume}{45}, \bibinfo{number}{1} (\bibinfo{date}{Oct.} \bibinfo{year}{2001}), \bibinfo{pages}{5–32}.
\newblock
\showISSN{1573-0565}
\href{https://doi.org/10.1023/A:1010933404324}{doi:\nolinkurl{10.1023/A:1010933404324}}


\bibitem[CDC(2024)]%
        {CDC_2024}
\bibfield{author}{\bibinfo{person}{CDC}.} \bibinfo{year}{2024}\natexlab{}.
\newblock \bibinfo{title}{Centers for Disease Control and Prevention}.
\newblock
\urldef\tempurl%
\url{https://www.cdc.gov/index.html}
\showURL{%
\tempurl}


\bibitem[Chen and Guestrin(2016)]%
        {Chen_Guestrin_2016}
\bibfield{author}{\bibinfo{person}{Tianqi Chen} {and} \bibinfo{person}{Carlos Guestrin}.} \bibinfo{year}{2016}\natexlab{}.
\newblock \showarticletitle{XGBoost: A Scalable Tree Boosting System}. In \bibinfo{booktitle}{\emph{Proceedings of the 22nd ACM SIGKDD International Conference on Knowledge Discovery and Data Mining}}. \bibinfo{pages}{785–794}.
\newblock
\href{https://doi.org/10.1145/2939672.2939785}{doi:\nolinkurl{10.1145/2939672.2939785}}
\newblock
\shownote{arXiv:1603.02754 [cs]}.


\bibitem[Christin et~al\mbox{.}(2019)]%
        {Christin_Hervet_Lecomte_2019}
\bibfield{author}{\bibinfo{person}{Sylvain Christin}, \bibinfo{person}{Éric Hervet}, {and} \bibinfo{person}{Nicolas Lecomte}.} \bibinfo{year}{2019}\natexlab{}.
\newblock \showarticletitle{Applications for deep learning in ecology}.
\newblock \bibinfo{journal}{\emph{Methods in Ecology and Evolution}} \bibinfo{volume}{10}, \bibinfo{number}{10} (\bibinfo{year}{2019}), \bibinfo{pages}{1632–1644}.
\newblock
\showISSN{2041-210X}
\href{https://doi.org/10.1111/2041-210X.13256}{doi:\nolinkurl{10.1111/2041-210X.13256}}


\bibitem[{Colorado Department of Public Health and Environment}(nd)]%
        {CDPHE_PFAS_Action_Plan}
\bibfield{author}{\bibinfo{person}{{Colorado Department of Public Health and Environment}}.} \bibinfo{year}{n.d.}\natexlab{}.
\newblock \bibinfo{title}{PFAS Action Plan}.
\newblock
\urldef\tempurl%
\url{https://cdphe.colorado.gov/chemicals-from-toxic-firefighting-foam-pfas/pfas-action-plan}
\showURL{%
\tempurl}


\bibitem[Cong et~al\mbox{.}(2023)]%
        {Cong_Khanna_Meng_Liu_Rozi_He_Burke_Lobell_Ermon_2023}
\bibfield{author}{\bibinfo{person}{Yezhen Cong}, \bibinfo{person}{Samar Khanna}, \bibinfo{person}{Chenlin Meng}, \bibinfo{person}{Patrick Liu}, \bibinfo{person}{Erik Rozi}, \bibinfo{person}{Yutong He}, \bibinfo{person}{Marshall Burke}, \bibinfo{person}{David~B. Lobell}, {and} \bibinfo{person}{Stefano Ermon}.} \bibinfo{year}{2023}\natexlab{}.
\newblock \showarticletitle{SatMAE: Pre-training Transformers for Temporal and Multi-Spectral Satellite Imagery}.
\newblock  \bibinfo{number}{arXiv:2207.08051} (\bibinfo{date}{Jan.} \bibinfo{year}{2023}).
\newblock
\href{https://doi.org/10.48550/arXiv.2207.08051}{doi:\nolinkurl{10.48550/arXiv.2207.08051}}
\newblock
\shownote{arXiv:2207.08051 [cs]}.


\bibitem[Crone et~al\mbox{.}(2019)]%
        {Crone_Speth_Wahman_Smith_Abulikemu_Kleiner_Pressman_2019}
\bibfield{author}{\bibinfo{person}{Brian~C. Crone}, \bibinfo{person}{Thomas~F. Speth}, \bibinfo{person}{David~G. Wahman}, \bibinfo{person}{Samantha~J. Smith}, \bibinfo{person}{Gulizhaer Abulikemu}, \bibinfo{person}{Eric~J. Kleiner}, {and} \bibinfo{person}{Jonathan~G. Pressman}.} \bibinfo{year}{2019}\natexlab{}.
\newblock \showarticletitle{Occurrence of Per- and Polyfluoroalkyl Substances (PFAS) in Source Water and Their Treatment in Drinking Water}.
\newblock \bibinfo{journal}{\emph{Critical reviews in environmental science and technology}} \bibinfo{volume}{49}, \bibinfo{number}{24} (\bibinfo{date}{June} \bibinfo{year}{2019}), \bibinfo{pages}{2359–2396}.
\newblock
\showISSN{1064-3389}
\href{https://doi.org/10.1080/10643389.2019.1614848}{doi:\nolinkurl{10.1080/10643389.2019.1614848}}


\bibitem[DeLuca et~al\mbox{.}(2023)]%
        {DeLuca_Mullikin_Brumm_2023}
\bibfield{author}{\bibinfo{person}{Nicole~M. DeLuca}, \bibinfo{person}{Ashley Mullikin}, \bibinfo{person}{Peter Brumm}, \bibinfo{person}{Ana~G. Rappold}, {and} \bibinfo{person}{Elaine Cohen~Hubal}.} \bibinfo{year}{2023}\natexlab{}.
\newblock \showarticletitle{Using Geospatial Data and Random Forest To Predict PFAS Contamination in Fish Tissue in the Columbia River Basin, United States}.
\newblock \bibinfo{journal}{\emph{Environmental Science \& Technology}} \bibinfo{volume}{57}, \bibinfo{number}{37} (\bibinfo{date}{sep} \bibinfo{year}{2023}), \bibinfo{pages}{14024–14035}.
\newblock
\showISSN{0013-936X}
\href{https://doi.org/10.1021/acs.est.3c03670}{doi:\nolinkurl{10.1021/acs.est.3c03670}}


\bibitem[Doudrick(2024)]%
        {Doudrick_2024}
\bibfield{author}{\bibinfo{person}{Kyle Doudrick}.} \bibinfo{year}{2024}\natexlab{}.
\newblock \bibinfo{title}{Removing PFAS from public water will cost billions and take time • Virginia Mercury}.
\newblock
\urldef\tempurl%
\url{https://virginiamercury.com/2024/04/18/removing-pfas-from-public-water-will-cost-billions-and-take-time/}
\showURL{%
\tempurl}


\bibitem[{EcoCenter}(nd)]%
        {ProtectingCommunitiesFromPFAS}
\bibfield{author}{\bibinfo{person}{{EcoCenter}}.} \bibinfo{year}{n.d.}\natexlab{}.
\newblock \bibinfo{title}{Protecting Communities from PFAS Pollution}.
\newblock
\urldef\tempurl%
\url{https://www.ecocenter.org/our-work/protecting-communities-pfas-pollution}
\showURL{%
\tempurl}
\newblock
\shownote{PFAS contamination is one of the most pressing environmental health issues we face. These chemicals impact the lives and health of millions of Americans. Per- and polyfluoroalkyl substances (PFAS) are a class of more than 12,000 different chemicals with established links to adverse health impacts. There are more than 285 sites in Michigan that have been identified as PFAS contaminated, and in the United States that number is over 6,000.}.


\bibitem[{Environmental Working Group}(2023a)]%
        {EWGstudy2023}
\bibfield{author}{\bibinfo{person}{{Environmental Working Group}}.} \bibinfo{year}{2023}\natexlab{a}.
\newblock \bibinfo{title}{EWG Study: Eating One Freshwater Fish Equals a Month of Drinking Water Laced with PFOS}.
\newblock
\urldef\tempurl%
\url{https://www.ewg.org/news-insights/news-release/2023/01/ewg-study-eating-one-freshwater-fish-equals-month-drinking}
\showURL{%
\tempurl}
\newblock
\shownote{A new study by Environmental Working Group scientists finds that consumption of just a single serving of freshwater fish per year could be equal to a month of drinking water laced with the “forever chemical” PFOS at high levels that may be harmful.}.


\bibitem[{Environmental Working Group}(2023b)]%
        {ewg2023foreverfish}
\bibfield{author}{\bibinfo{person}{{Environmental Working Group}}.} \bibinfo{year}{2023}\natexlab{b}.
\newblock \bibinfo{title}{Forever Chemicals in Freshwater Fish: Mapping a Growing Environmental Justice Crisis}.
\newblock
\urldef\tempurl%
\url{https://www.ewg.org/news-insights/news/2023/01/forever-chemicals-freshwater-fish-mapping-growing-environmental-justice}
\showURL{%
\tempurl}
\newblock
\shownote{Accessed: 2025-07-27}.


\bibitem[{Environmental Working Group}(2024)]%
        {EWG}
\bibfield{author}{\bibinfo{person}{{Environmental Working Group}}.} \bibinfo{year}{2024}\natexlab{}.
\newblock \bibinfo{title}{Interactive Map: PFAS Contamination Crisis: New Data Show 8,865 Sites in 50 States}.
\newblock
\urldef\tempurl%
\url{http://www.ewg.org/interactive-maps/pfas_contamination/map/}
\showURL{%
\tempurl}


\bibitem[{Esri}(2024)]%
        {ArcGIS_FlowDirection}
\bibfield{author}{\bibinfo{person}{{Esri}}.} \bibinfo{year}{2024}\natexlab{}.
\newblock \bibinfo{title}{Flow Direction (Spatial Analyst) --- ArcGIS Pro Documentation}.
\newblock \bibinfo{howpublished}{\url{https://pro.arcgis.com/en/pro-app/latest/tool-reference/spatial-analyst/flow-direction.htm}}.
\newblock
\newblock
\shownote{Accessed: 2025-07-17}.


\bibitem[{Esri}(nd)]%
        {esri_kriging}
\bibfield{author}{\bibinfo{person}{{Esri}}.} \bibinfo{year}{n.d.}\natexlab{}.
\newblock \bibinfo{title}{How Kriging Works — ArcGIS Pro Documentation}.
\newblock
\urldef\tempurl%
\url{https://pro.arcgis.com/en/pro-app/latest/tool-reference/3d-analyst/how-kriging-works.htm}
\showURL{%
\tempurl}
\newblock
\shownote{Accessed: 2025-07-27}.


\bibitem[GISGeography(2017)]%
        {GISGeography_2017}
\bibfield{author}{\bibinfo{person}{GISGeography}.} \bibinfo{year}{2017}\natexlab{}.
\newblock \bibinfo{title}{Kriging Interpolation - The Prediction Is Strong in this One}.
\newblock
\urldef\tempurl%
\url{https://gisgeography.com/kriging-interpolation-prediction/}
\showURL{%
\tempurl}


\bibitem[Gokcesu and Gokcesu(2021)]%
        {Gokcesu_Gokcesu_2021}
\bibfield{author}{\bibinfo{person}{Kaan Gokcesu} {and} \bibinfo{person}{Hakan Gokcesu}.} \bibinfo{year}{2021}\natexlab{}.
\newblock \showarticletitle{Generalized Huber Loss for Robust Learning and its Efficient Minimization for a Robust Statistics}.
\newblock  \bibinfo{number}{arXiv:2108.12627} (\bibinfo{date}{Aug.} \bibinfo{year}{2021}).
\newblock
\href{https://doi.org/10.48550/arXiv.2108.12627}{doi:\nolinkurl{10.48550/arXiv.2108.12627}}
\newblock
\shownote{arXiv:2108.12627 [stat]}.


\bibitem[Goldberger and Ben-Reuven(2017)]%
        {Goldberger_Ben-Reuven_2017}
\bibfield{author}{\bibinfo{person}{Jacob Goldberger} {and} \bibinfo{person}{Ehud Ben-Reuven}.} \bibinfo{year}{2017}\natexlab{}.
\newblock \showarticletitle{Training deep neural-networks using a noise adaptation layer}.
\newblock
\urldef\tempurl%
\url{https://openreview.net/forum?id=H12GRgcxg}
\showURL{%
\tempurl}


\bibitem[Guelfo et~al\mbox{.}(2021)]%
        {Guelfo_2021}
\bibfield{author}{\bibinfo{person}{Jennifer~L. Guelfo}, \bibinfo{person}{Stephen Korzeniowski}, \bibinfo{person}{Marc~A. Mills}, \bibinfo{person}{Janet Anderson}, \bibinfo{person}{Richard~H. Anderson}, \bibinfo{person}{Jennifer~A. Arblaster}, \bibinfo{person}{Jason~M. Conder}, \bibinfo{person}{Ian~T. Cousins}, \bibinfo{person}{Kavitha Dasu}, \bibinfo{person}{Barbara~J. Henry}, \bibinfo{person}{Linda~S. Lee}, \bibinfo{person}{Jinxia Liu}, \bibinfo{person}{Erica~R. McKenzie}, {and} \bibinfo{person}{Janice Willey}.} \bibinfo{year}{2021}\natexlab{}.
\newblock \showarticletitle{Environmental Sources, Chemistry, Fate, and Transport of Per‐ and Polyfluoroalkyl Substances: State of the Science, Key Knowledge Gaps, and Recommendations Presented at the August 2019 SETAC Focus Topic Meeting}.
\newblock \bibinfo{journal}{\emph{Environmental Toxicology and Chemistry}} \bibinfo{volume}{40}, \bibinfo{number}{12} (\bibinfo{date}{Dec.} \bibinfo{year}{2021}), \bibinfo{pages}{3234--3260}.
\newblock
\showISSN{0730-7268}
\href{https://doi.org/10.1002/etc.5182}{doi:\nolinkurl{10.1002/etc.5182}}


\bibitem[Guo et~al\mbox{.}(2017)]%
        {Guo_Pleiss_Sun_Weinberger_2017}
\bibfield{author}{\bibinfo{person}{Chuan Guo}, \bibinfo{person}{Geoff Pleiss}, \bibinfo{person}{Yu Sun}, {and} \bibinfo{person}{Kilian~Q. Weinberger}.} \bibinfo{year}{2017}\natexlab{}.
\newblock \showarticletitle{On Calibration of Modern Neural Networks}.
\newblock  \bibinfo{number}{arXiv:1706.04599} (\bibinfo{date}{Aug.} \bibinfo{year}{2017}).
\newblock
\href{https://doi.org/10.48550/arXiv.1706.04599}{doi:\nolinkurl{10.48550/arXiv.1706.04599}}
\newblock
\shownote{arXiv:1706.04599 [cs]}.


\bibitem[Hua et~al\mbox{.}(2022)]%
        {Hua_Marcos_Mou_Zhu_Tuia_2022}
\bibfield{author}{\bibinfo{person}{Yuansheng Hua}, \bibinfo{person}{Diego Marcos}, \bibinfo{person}{Lichao Mou}, \bibinfo{person}{Xiao~Xiang Zhu}, {and} \bibinfo{person}{Devis Tuia}.} \bibinfo{year}{2022}\natexlab{}.
\newblock \showarticletitle{Semantic Segmentation of Remote Sensing Images with Sparse Annotations}.
\newblock \bibinfo{journal}{\emph{IEEE Geoscience and Remote Sensing Letters}}  \bibinfo{volume}{19} (\bibinfo{year}{2022}), \bibinfo{pages}{1–5}.
\newblock
\showISSN{1545-598X, 1558-0571}
\href{https://doi.org/10.1109/LGRS.2021.3051053}{doi:\nolinkurl{10.1109/LGRS.2021.3051053}}
\newblock
\shownote{arXiv:2101.03492 [cs]}.


\bibitem[Karniadakis et~al\mbox{.}(2021)]%
        {Karniadakis_Kevrekidis_Lu_Perdikaris_Wang_Yang_2021}
\bibfield{author}{\bibinfo{person}{George~Em Karniadakis}, \bibinfo{person}{Ioannis~G. Kevrekidis}, \bibinfo{person}{Lu Lu}, \bibinfo{person}{Paris Perdikaris}, \bibinfo{person}{Sifan Wang}, {and} \bibinfo{person}{Liu Yang}.} \bibinfo{year}{2021}\natexlab{}.
\newblock \showarticletitle{Physics-informed machine learning}.
\newblock \bibinfo{journal}{\emph{Nature Reviews Physics}} \bibinfo{volume}{3}, \bibinfo{number}{6} (\bibinfo{date}{June} \bibinfo{year}{2021}), \bibinfo{pages}{422–440}.
\newblock
\showISSN{2522-5820}
\href{https://doi.org/10.1038/s42254-021-00314-5}{doi:\nolinkurl{10.1038/s42254-021-00314-5}}


\bibitem[Kerner et~al\mbox{.}(2024)]%
        {Kerner_Nakalembe_Yang_Zvonkov_McWeeny_Tseng_Becker-Reshef_2024}
\bibfield{author}{\bibinfo{person}{Hannah Kerner}, \bibinfo{person}{Catherine Nakalembe}, \bibinfo{person}{Adam Yang}, \bibinfo{person}{Ivan Zvonkov}, \bibinfo{person}{Ryan McWeeny}, \bibinfo{person}{Gabriel Tseng}, {and} \bibinfo{person}{Inbal Becker-Reshef}.} \bibinfo{year}{2024}\natexlab{}.
\newblock \showarticletitle{How accurate are existing land cover maps for agriculture in Sub-Saharan Africa?}
\newblock \bibinfo{journal}{\emph{Scientific Data}} \bibinfo{volume}{11}, \bibinfo{number}{1} (\bibinfo{date}{May} \bibinfo{year}{2024}), \bibinfo{pages}{486}.
\newblock
\showISSN{2052-4463}
\href{https://doi.org/10.1038/s41597-024-03306-z}{doi:\nolinkurl{10.1038/s41597-024-03306-z}}


\bibitem[Langenbach and Wilson(2021)]%
        {Langenbach_Wilson_2021}
\bibfield{author}{\bibinfo{person}{Blake Langenbach} {and} \bibinfo{person}{Mark Wilson}.} \bibinfo{year}{2021}\natexlab{}.
\newblock \showarticletitle{Per- and Polyfluoroalkyl Substances (PFAS): Significance and Considerations within the Regulatory Framework of the USA}.
\newblock \bibinfo{journal}{\emph{International Journal of Environmental Research and Public Health}} \bibinfo{volume}{18}, \bibinfo{number}{21} (\bibinfo{date}{Oct.} \bibinfo{year}{2021}), \bibinfo{pages}{11142}.
\newblock
\showISSN{1660-4601}
\href{https://doi.org/10.3390/ijerph182111142}{doi:\nolinkurl{10.3390/ijerph182111142}}


\bibitem[Largueche(2006)]%
        {Largueche_2006}
\bibfield{author}{\bibinfo{person}{Fatima-Zohra~Benmost Largueche}.} \bibinfo{year}{2006}\natexlab{}.
\newblock \showarticletitle{Estimating Soil Contamination with Kriging Interpolation Method}.
\newblock \bibinfo{journal}{\emph{American Journal of Applied Sciences}} \bibinfo{volume}{3}, \bibinfo{number}{6} (\bibinfo{date}{June} \bibinfo{year}{2006}), \bibinfo{pages}{1894–1898}.
\newblock
\showISSN{15469239}
\href{https://doi.org/10.3844/ajassp.2006.1894.1898}{doi:\nolinkurl{10.3844/ajassp.2006.1894.1898}}


\bibitem[Lechner et~al\mbox{.}(2009)]%
        {Lechner_Stein_Jones_Ferwerda_2009}
\bibfield{author}{\bibinfo{person}{Alex~Mark Lechner}, \bibinfo{person}{Alfred Stein}, \bibinfo{person}{Simon~D. Jones}, {and} \bibinfo{person}{Jelle~Garke Ferwerda}.} \bibinfo{year}{2009}\natexlab{}.
\newblock \showarticletitle{Remote sensing of small and linear features: quantifying the effects of patch size and length, grid position and detectability on land cover mapping}.
\newblock \bibinfo{journal}{\emph{Remote Sensing of Environment}} \bibinfo{volume}{113}, \bibinfo{number}{10} (\bibinfo{date}{Oct.} \bibinfo{year}{2009}), \bibinfo{pages}{2194–2204}.
\newblock
\showISSN{0034-4257}
\href{https://doi.org/10.1016/j.rse.2009.06.002}{doi:\nolinkurl{10.1016/j.rse.2009.06.002}}


\bibitem[Li et~al\mbox{.}(2020)]%
        {Li_Kovachki_Azizzadenesheli_Liu_Bhattacharya_Stuart_Anandkumar_2020}
\bibfield{author}{\bibinfo{person}{Zongyi Li}, \bibinfo{person}{Nikola Kovachki}, \bibinfo{person}{Kamyar Azizzadenesheli}, \bibinfo{person}{Burigede Liu}, \bibinfo{person}{Kaushik Bhattacharya}, \bibinfo{person}{Andrew Stuart}, {and} \bibinfo{person}{Anima Anandkumar}.} \bibinfo{year}{2020}\natexlab{}.
\newblock \bibinfo{title}{Fourier Neural Operator for Parametric Partial Differential Equations}.
\newblock
\urldef\tempurl%
\url{https://arxiv.org/abs/2010.08895v3}
\showURL{%
\tempurl}


\bibitem[Liddie et~al\mbox{.}(2024)]%
        {Liddie_Bind_Karra_Sunderland_2024}
\bibfield{author}{\bibinfo{person}{Jahred~M. Liddie}, \bibinfo{person}{Marie-Abèle Bind}, \bibinfo{person}{Mahesh Karra}, {and} \bibinfo{person}{Elsie~M. Sunderland}.} \bibinfo{year}{2024}\natexlab{}.
\newblock \showarticletitle{County-level associations between drinking water PFAS contamination and COVID-19 mortality in the United States}.
\newblock \bibinfo{journal}{\emph{Journal of Exposure Science \& Environmental Epidemiology}} (\bibinfo{date}{Oct.} \bibinfo{year}{2024}), \bibinfo{pages}{1–8}.
\newblock
\showISSN{1559-064X}
\href{https://doi.org/10.1038/s41370-024-00723-5}{doi:\nolinkurl{10.1038/s41370-024-00723-5}}


\bibitem[Lin et~al\mbox{.}(2018)]%
        {Lin_Goyal_Girshick_He_Dollár_2018}
\bibfield{author}{\bibinfo{person}{Tsung-Yi Lin}, \bibinfo{person}{Priya Goyal}, \bibinfo{person}{Ross Girshick}, \bibinfo{person}{Kaiming He}, {and} \bibinfo{person}{Piotr Dollár}.} \bibinfo{year}{2018}\natexlab{}.
\newblock \showarticletitle{Focal Loss for Dense Object Detection}.
\newblock  \bibinfo{number}{arXiv:1708.02002} (\bibinfo{date}{Feb.} \bibinfo{year}{2018}).
\newblock
\href{https://doi.org/10.48550/arXiv.1708.02002}{doi:\nolinkurl{10.48550/arXiv.1708.02002}}
\newblock
\shownote{arXiv:1708.02002 [cs]}.


\bibitem[Loshchilov and Hutter(2019)]%
        {Loshchilov_Hutter_2019}
\bibfield{author}{\bibinfo{person}{Ilya Loshchilov} {and} \bibinfo{person}{Frank Hutter}.} \bibinfo{year}{2019}\natexlab{}.
\newblock \showarticletitle{Decoupled Weight Decay Regularization}.
\newblock  \bibinfo{number}{arXiv:1711.05101} (\bibinfo{date}{Jan.} \bibinfo{year}{2019}).
\newblock
\href{https://doi.org/10.48550/arXiv.1711.05101}{doi:\nolinkurl{10.48550/arXiv.1711.05101}}
\newblock
\shownote{arXiv:1711.05101 [cs]}.


\bibitem[Manz(2024)]%
        {Manz_2024}
\bibfield{author}{\bibinfo{person}{Katherine~E. Manz}.} \bibinfo{year}{2024}\natexlab{}.
\newblock \showarticletitle{Considerations for Measurements of Aggregate PFAS Exposure in Precision Environmental Health}.
\newblock \bibinfo{journal}{\emph{ACS Measurement Science Au}} \bibinfo{volume}{4}, \bibinfo{number}{6} (\bibinfo{date}{Dec.} \bibinfo{year}{2024}), \bibinfo{pages}{620–628}.
\newblock
\href{https://doi.org/10.1021/acsmeasuresciau.4c00052}{doi:\nolinkurl{10.1021/acsmeasuresciau.4c00052}}


\bibitem[{Michigan Department of Environment, Great Lakes, and Energy}(2023)]%
        {egle_pfas_surface_water}
\bibfield{author}{\bibinfo{person}{{Michigan Department of Environment, Great Lakes, and Energy}}.} \bibinfo{year}{2023}\natexlab{}.
\newblock \bibinfo{title}{PFAS Surface Water Sampling}.
\newblock
\urldef\tempurl%
\url{https://gis-egle.hub.arcgis.com/datasets/391cca4f364845829abcd5a92093c631_1/about}
\showURL{%
\tempurl}
\newblock
\shownote{Accessed: 2025-07-27}.


\bibitem[{Michigan Department of Environment, Great Lakes, and Energy}(2025)]%
        {egle_fcmp_2025}
\bibfield{author}{\bibinfo{person}{{Michigan Department of Environment, Great Lakes, and Energy}}.} \bibinfo{year}{2025}\natexlab{}.
\newblock \bibinfo{title}{Michigan Fish Contaminant Monitoring Sampling Sites and Select Results (Consolidated)}.
\newblock
\urldef\tempurl%
\url{https://gis-egle.hub.arcgis.com/maps/d4bbb519842c44638eeb9a15461c441f/about}
\showURL{%
\tempurl}
\newblock
\shownote{Accessed: 2025-07-27}.


\bibitem[{Michigan Department of Health and Human Services}(nd)]%
        {MDHHS_FindYourArea_SafeFish}
\bibfield{author}{\bibinfo{person}{{Michigan Department of Health and Human Services}}.} \bibinfo{year}{n.d.}\natexlab{}.
\newblock \bibinfo{title}{Find Your Area -- Eat Safe Fish Guidelines}.
\newblock \bibinfo{howpublished}{\url{https://www.michigan.gov/mdhhs/safety-injury-prev/environmental-health/topics/eatsafefish/find-your-area}}.
\newblock
\newblock
\shownote{Accessed: 2026-02-05}.


\bibitem[Natarajan et~al\mbox{.}(2013)]%
        {Natarajan_Dhillon_Ravikumar_Tewari_2013}
\bibfield{author}{\bibinfo{person}{Nagarajan Natarajan}, \bibinfo{person}{Inderjit~S Dhillon}, \bibinfo{person}{Pradeep~K Ravikumar}, {and} \bibinfo{person}{Ambuj Tewari}.} \bibinfo{year}{2013}\natexlab{}.
\newblock \showarticletitle{Learning with Noisy Labels}. In \bibinfo{booktitle}{\emph{Advances in Neural Information Processing Systems}}, Vol.~\bibinfo{volume}{26}. \bibinfo{publisher}{Curran Associates, Inc.}
\newblock
\urldef\tempurl%
\url{https://papers.nips.cc/paper_files/paper/2013/hash/3871bd64012152bfb53fdf04b401193f-Abstract.html}
\showURL{%
\tempurl}


\bibitem[{National Institute of Environmental Health Sciences}(nd)]%
        {niehs_pfas}
\bibfield{author}{\bibinfo{person}{{National Institute of Environmental Health Sciences}}.} \bibinfo{year}{n.d.}\natexlab{}.
\newblock \bibinfo{title}{Perfluoroalkyl and Polyfluoroalkyl Substances (PFAS)}.
\newblock \bibinfo{howpublished}{\url{https://www.niehs.nih.gov/health/topics/agents/pfc}}.
\newblock
\urldef\tempurl%
\url{https://www.niehs.nih.gov/health/topics/agents/pfc}
\showURL{%
\tempurl}
\newblock
\shownote{Accessed: 2025-07-27}.


\bibitem[Northcutt et~al\mbox{.}(2021)]%
        {Northcutt_Jiang_Chuang_2021}
\bibfield{author}{\bibinfo{person}{Curtis Northcutt}, \bibinfo{person}{Lu Jiang}, {and} \bibinfo{person}{Isaac Chuang}.} \bibinfo{year}{2021}\natexlab{}.
\newblock \showarticletitle{Confident Learning: Estimating Uncertainty in Dataset Labels}.
\newblock \bibinfo{journal}{\emph{J. Artif. Int. Res.}}  \bibinfo{volume}{70} (\bibinfo{date}{May} \bibinfo{year}{2021}), \bibinfo{pages}{1373–1411}.
\newblock
\showISSN{1076-9757}
\href{https://doi.org/10.1613/jair.1.12125}{doi:\nolinkurl{10.1613/jair.1.12125}}


\bibitem[Pelch et~al\mbox{.}(2022)]%
        {Pelch_Reade_Kwiatkowski_Merced-Nieves_Cavalier_Schultz_Wolffe_Varshavsky_2022}
\bibfield{author}{\bibinfo{person}{Katherine~E. Pelch}, \bibinfo{person}{Anna Reade}, \bibinfo{person}{Carol~F. Kwiatkowski}, \bibinfo{person}{Francheska~M. Merced-Nieves}, \bibinfo{person}{Haleigh Cavalier}, \bibinfo{person}{Kim Schultz}, \bibinfo{person}{Taylor Wolffe}, {and} \bibinfo{person}{Julia Varshavsky}.} \bibinfo{year}{2022}\natexlab{}.
\newblock \showarticletitle{The PFAS-Tox Database: A systematic evidence map of health studies on 29 per- and polyfluoroalkyl substances}.
\newblock \bibinfo{journal}{\emph{Environment International}}  \bibinfo{volume}{167} (\bibinfo{date}{Sept.} \bibinfo{year}{2022}), \bibinfo{pages}{107408}.
\newblock
\showISSN{0160-4120}
\href{https://doi.org/10.1016/j.envint.2022.107408}{doi:\nolinkurl{10.1016/j.envint.2022.107408}}


\bibitem[Reed et~al\mbox{.}(2015)]%
        {Reed_Lee_Anguelov_Szegedy_Erhan_Rabinovich_2015}
\bibfield{author}{\bibinfo{person}{Scott Reed}, \bibinfo{person}{Honglak Lee}, \bibinfo{person}{Dragomir Anguelov}, \bibinfo{person}{Christian Szegedy}, \bibinfo{person}{Dumitru Erhan}, {and} \bibinfo{person}{Andrew Rabinovich}.} \bibinfo{year}{2015}\natexlab{}.
\newblock \showarticletitle{Training Deep Neural Networks on Noisy Labels with Bootstrapping}.
\newblock  \bibinfo{number}{arXiv:1412.6596} (\bibinfo{date}{April} \bibinfo{year}{2015}).
\newblock
\href{https://doi.org/10.48550/arXiv.1412.6596}{doi:\nolinkurl{10.48550/arXiv.1412.6596}}
\newblock
\shownote{arXiv:1412.6596 [cs]}.


\bibitem[Reichstein et~al\mbox{.}(2019)]%
        {Reichstein_Camps-Valls_Stevens_Jung_Denzler_Carvalhais_Prabhat_2019}
\bibfield{author}{\bibinfo{person}{Markus Reichstein}, \bibinfo{person}{Gustau Camps-Valls}, \bibinfo{person}{Bjorn Stevens}, \bibinfo{person}{Martin Jung}, \bibinfo{person}{Joachim Denzler}, \bibinfo{person}{Nuno Carvalhais}, {and} \bibinfo{person}{Prabhat}.} \bibinfo{year}{2019}\natexlab{}.
\newblock \showarticletitle{Deep learning and process understanding for data-driven Earth system science}.
\newblock \bibinfo{journal}{\emph{Nature}} \bibinfo{volume}{566}, \bibinfo{number}{7743} (\bibinfo{date}{Feb.} \bibinfo{year}{2019}), \bibinfo{pages}{195–204}.
\newblock
\showISSN{1476-4687}
\href{https://doi.org/10.1038/s41586-019-0912-1}{doi:\nolinkurl{10.1038/s41586-019-0912-1}}


\bibitem[Robinson et~al\mbox{.}(2020)]%
        {Robinson_Ortiz_Malkin_Elias_Peng_Morris_Dilkina_Jojic_2020}
\bibfield{author}{\bibinfo{person}{Caleb Robinson}, \bibinfo{person}{Anthony Ortiz}, \bibinfo{person}{Kolya Malkin}, \bibinfo{person}{Blake Elias}, \bibinfo{person}{Andi Peng}, \bibinfo{person}{Dan Morris}, \bibinfo{person}{Bistra Dilkina}, {and} \bibinfo{person}{Nebojsa Jojic}.} \bibinfo{year}{2020}\natexlab{}.
\newblock \showarticletitle{Human-Machine Collaboration for Fast Land Cover Mapping}.
\newblock \bibinfo{journal}{\emph{Proceedings of the AAAI Conference on Artificial Intelligence}} \bibinfo{volume}{34}, \bibinfo{number}{0303} (\bibinfo{date}{April} \bibinfo{year}{2020}), \bibinfo{pages}{2509–2517}.
\newblock
\showISSN{2374-3468}
\href{https://doi.org/10.1609/aaai.v34i03.5633}{doi:\nolinkurl{10.1609/aaai.v34i03.5633}}


\bibitem[Russo et~al\mbox{.}(2020)]%
        {Russo_2020}
\bibfield{author}{\bibinfo{person}{Stefania Russo}, \bibinfo{person}{Moritz Lürig}, \bibinfo{person}{Wenjin Hao}, \bibinfo{person}{Blake Matthews}, {and} \bibinfo{person}{Kris Villez}.} \bibinfo{year}{2020}\natexlab{}.
\newblock \showarticletitle{Active Learning for Anomaly Detection in Environmental Data}.
\newblock \bibinfo{journal}{\emph{Environmental Modelling \& Software}}  \bibinfo{volume}{134} (\bibinfo{date}{Dec.} \bibinfo{year}{2020}), \bibinfo{pages}{104869}.
\newblock
\showISSN{1364-8152}
\href{https://doi.org/10.1016/j.envsoft.2020.104869}{doi:\nolinkurl{10.1016/j.envsoft.2020.104869}}


\bibitem[Safonova et~al\mbox{.}(2023)]%
        {Safonova_2023}
\bibfield{author}{\bibinfo{person}{Anastasiia Safonova}, \bibinfo{person}{Gohar Ghazaryan}, \bibinfo{person}{Stefan Stiller}, \bibinfo{person}{Magdalena Main-Knorn}, \bibinfo{person}{Claas Nendel}, {and} \bibinfo{person}{Masahiro Ryo}.} \bibinfo{year}{2023}\natexlab{}.
\newblock \showarticletitle{Ten Deep Learning Techniques to Address Small Data Problems with Remote Sensing}.
\newblock \bibinfo{journal}{\emph{International Journal of Applied Earth Observation and Geoinformation}}  \bibinfo{volume}{125} (\bibinfo{date}{Dec.} \bibinfo{year}{2023}), \bibinfo{pages}{103569}.
\newblock
\showISSN{1569-8432}
\href{https://doi.org/10.1016/j.jag.2023.103569}{doi:\nolinkurl{10.1016/j.jag.2023.103569}}


\bibitem[Salvatore et~al\mbox{.}(2022)]%
        {Salvatore_Mok_Garrett_2022}
\bibfield{author}{\bibinfo{person}{Derrick Salvatore}, \bibinfo{person}{Kira Mok}, \bibinfo{person}{Kimberly~K. Garrett}, \bibinfo{person}{Grace Poudrier}, \bibinfo{person}{Phil Brown}, \bibinfo{person}{Linda~S. Birnbaum}, \bibinfo{person}{Gretta Goldenman}, \bibinfo{person}{Mark~F. Miller}, \bibinfo{person}{Sharyle Patton}, \bibinfo{person}{Maddy Poehlein}, \bibinfo{person}{Julia Varshavsky}, {and} \bibinfo{person}{Alissa Cordner}.} \bibinfo{year}{2022}\natexlab{}.
\newblock \showarticletitle{Presumptive Contamination: A New Approach to PFAS Contamination Based on Likely Sources}.
\newblock \bibinfo{journal}{\emph{Environmental Science \& Technology Letters}} \bibinfo{volume}{9}, \bibinfo{number}{11} (\bibinfo{date}{nov} \bibinfo{year}{2022}), \bibinfo{pages}{983–990}.
\newblock
\showISSN{2328-8930}
\href{https://doi.org/10.1021/acs.estlett.2c00502}{doi:\nolinkurl{10.1021/acs.estlett.2c00502}}


\bibitem[Schroeder et~al\mbox{.}(2021)]%
        {Schroeder_Bond_Foley_2021}
\bibfield{author}{\bibinfo{person}{Tim Schroeder}, \bibinfo{person}{David Bond}, {and} \bibinfo{person}{Janet Foley}.} \bibinfo{year}{2021}\natexlab{}.
\newblock \showarticletitle{PFAS soil and groundwater contamination via industrial airborne emission and land deposition in SW Vermont and Eastern New York State, USA}.
\newblock \bibinfo{journal}{\emph{Environmental Science. Processes \& Impacts}} \bibinfo{volume}{23}, \bibinfo{number}{2} (\bibinfo{date}{March} \bibinfo{year}{2021}), \bibinfo{pages}{291–301}.
\newblock
\showISSN{2050-7895}
\href{https://doi.org/10.1039/d0em00427h}{doi:\nolinkurl{10.1039/d0em00427h}}


\bibitem[{SciPy Developers}(2019)]%
        {scipy_distance_transform_edt}
\bibfield{author}{\bibinfo{person}{{SciPy Developers}}.} \bibinfo{year}{2019}\natexlab{}.
\newblock \bibinfo{title}{distance\_transform\_edt --- SciPy v1.16.0 Manual}.
\newblock \bibinfo{howpublished}{\url{https://docs.scipy.org/doc/scipy-1.16.0/reference/generated/scipy.ndimage.distance_transform_edt.html}}.
\newblock
\urldef\tempurl%
\url{https://docs.scipy.org/doc/scipy-1.16.0/reference/generated/scipy.ndimage.distance_transform_edt.html}
\showURL{%
\tempurl}
\newblock
\shownote{Accessed: 2025-07-27}.


\bibitem[Shoemaker and Tettenhorst(2020)]%
        {Shoemaker_2020_Method537}
\bibfield{author}{\bibinfo{person}{J. Shoemaker} {and} \bibinfo{person}{D. Tettenhorst}.} \bibinfo{year}{2020}\natexlab{}.
\newblock \bibinfo{title}{Method 537.1 Determination of Selected Per- and Polyfluorinated Alkyl Substances in Drinking Water by Solid Phase Extraction and Liquid Chromatography/Tandem Mass Spectrometry (LC/MS/MS)}.
\newblock
\urldef\tempurl%
\url{https://cfpub.epa.gov/si/si_public_record_report.cfm?dirEntryId=348508&Lab=CESER&simpleSearch=0&showCriteria=2&searchAll=537.1&TIMSType=&dateBeginPublishedPresented=03%2F24%2F2018}
\showURL{%
\tempurl}


\bibitem[{Sierra Club}(nd)]%
        {CommunityScience}
\bibfield{author}{\bibinfo{person}{{Sierra Club}}.} \bibinfo{year}{n.d.}\natexlab{}.
\newblock \bibinfo{title}{Community Science Testing for PFAS in Water or Biosolids}.
\newblock
\urldef\tempurl%
\url{https://www.sierraclub.org/grassroots-network/pfas/community-science-testing-pfas-water-or-biosolids}
\showURL{%
\tempurl}
\newblock
\shownote{Community science to test for PFAS in rivers and lakes and in sludge/biosolids}.


\bibitem[Singh and Verma(2019)]%
        {Singh_Verma_2019}
\bibfield{author}{\bibinfo{person}{Prafull Singh} {and} \bibinfo{person}{Pradipika Verma}.} \bibinfo{year}{2019}\natexlab{}.
\newblock \bibinfo{booktitle}{\emph{A Comparative Study of Spatial Interpolation Technique (IDW and Kriging) for Determining Groundwater Quality}}.
\newblock \bibinfo{publisher}{Elsevier}, \bibinfo{pages}{43–56}.
\newblock
\showISBNx{978-0-12-815413-7}
\href{https://doi.org/10.1016/B978-0-12-815413-7.00005-5}{doi:\nolinkurl{10.1016/B978-0-12-815413-7.00005-5}}


\bibitem[{SWAT}(nd)]%
        {SWAT2023}
\bibfield{author}{\bibinfo{person}{{SWAT}}.} \bibinfo{year}{n.d.}\natexlab{}.
\newblock \bibinfo{title}{Soil \& Water Assessment Tool}.
\newblock
\urldef\tempurl%
\url{https://swat.tamu.edu/}
\showURL{%
\tempurl}


\bibitem[Tokranov et~al\mbox{.}(2023)]%
        {Tokranov2023}
\bibfield{author}{\bibinfo{person}{Andrea~K. Tokranov}, \bibinfo{person}{Laura~M. Bexfield}, \bibinfo{person}{Katherine~M. Ransom}, \bibinfo{person}{James~A. Kingsbury}, \bibinfo{person}{Miranda~S. Fram}, \bibinfo{person}{Elise Watson}, \bibinfo{person}{Danielle~I. Dupuy}, \bibinfo{person}{Stefan Voss}, \bibinfo{person}{Bryant~C. Jurgens}, \bibinfo{person}{Paul~E. Stackelberg}, \bibinfo{person}{Delicia Beaty}, {and} \bibinfo{person}{Kelly Smalling}.} \bibinfo{year}{2023}\natexlab{}.
\newblock \bibinfo{title}{Predictions of PFAS in groundwater used as a source of drinking water and related data in the conterminous United States}.
\newblock
\href{https://doi.org/10.5066/P93RXTKJ}{doi:\nolinkurl{10.5066/P93RXTKJ}}
\newblock
\shownote{Creative Commons Zero v1.0 Universal}.


\bibitem[Torcoletti(2012)]%
        {ktorcoletti_2012}
\bibfield{author}{\bibinfo{person}{K. Torcoletti}.} \bibinfo{year}{2012}\natexlab{}.
\newblock \bibinfo{title}{What is MODFLOW?}
\newblock
\urldef\tempurl%
\url{https://www.waterloohydrogeologic.com/2012/06/22/what-is-modflow/}
\showURL{%
\tempurl}
\newblock
\shownote{Published by Waterloo Hydrogeologic. Accessed: 2025-07-27}.


\bibitem[{U.S. Environmental Protection Agency}(2023)]%
        {epa_hazard_index_2023}
\bibfield{author}{\bibinfo{person}{{U.S. Environmental Protection Agency}}.} \bibinfo{year}{2023}\natexlab{}.
\newblock \bibinfo{title}{How Do I Calculate the Hazard Index?}
\newblock
\urldef\tempurl%
\url{https://www.epa.gov/system/files/documents/2023-03/How%20do%20I%20calculate%20the%20Hazard%20Index._3.14.23.pdf}
\showURL{%
\tempurl}
\newblock
\shownote{EPA Fact Sheet, March 14}.


\bibitem[{U.S. Environmental Protection Agency}(nd)]%
        {US_EPA_Water_Pollution}
\bibfield{author}{\bibinfo{person}{{U.S. Environmental Protection Agency}}.} \bibinfo{year}{n.d.}\natexlab{}.
\newblock \bibinfo{title}{Water Pollution Search | ECHO | US EPA}.
\newblock
\urldef\tempurl%
\url{https://echo.epa.gov/trends/loading-tool/water-pollution-search}
\showURL{%
\tempurl}


\bibitem[{U.S. EPA}(2024)]%
        {Contaminants_to_Monitor_in_Fish_and_Shellfish_Advisory_Programs_2024}
\bibfield{author}{\bibinfo{person}{{U.S. EPA}}.} \bibinfo{year}{2024}\natexlab{}.
\newblock \showarticletitle{Contaminants to Monitor in Fish and Shellfish Advisory Programs: Compilation of Peer Review-Related Information}.
\newblock  (\bibinfo{year}{2024}).
\newblock


\bibitem[US~EPA(2024)]%
        {US_EPA_2024}
\bibfield{author}{\bibinfo{person}{OW US~EPA}.} \bibinfo{year}{2024}\natexlab{}.
\newblock \bibinfo{title}{2022 National Lakes Assessment - Fish Tissue Study}.
\newblock
\urldef\tempurl%
\url{https://www.epa.gov/choose-fish-and-shellfish-wisely/2022-national-lakes-assessment-fish-tissue-study}
\showURL{%
\tempurl}


\bibitem[{U.S. EPA, OW}(2015)]%
        {USEPA2015}
\bibfield{author}{\bibinfo{person}{{U.S. EPA, OW}}.} \bibinfo{year}{2015}\natexlab{}.
\newblock \bibinfo{title}{National Rivers and Streams Assessment}.
\newblock
\urldef\tempurl%
\url{https://www.epa.gov/national-aquatic-resource-surveys/nrsa}
\showURL{%
\tempurl}


\bibitem[{U.S. Geological Survey}(2023)]%
        {PFAS_tap}
\bibfield{author}{\bibinfo{person}{{U.S. Geological Survey}}.} \bibinfo{year}{2023}\natexlab{}.
\newblock
\urldef\tempurl%
\url{https://www.usgs.gov/tools/pfas-us-tapwater-interactive-dashboard}
\showURL{%
\tempurl}


\bibitem[{U.S. Geological Survey}(2024)]%
        {USGS_NLCD_2023}
\bibfield{author}{\bibinfo{person}{{U.S. Geological Survey}}.} \bibinfo{year}{2024}\natexlab{}.
\newblock \bibinfo{title}{National Land Cover Database | U.S. Geological Survey}.
\newblock
\urldef\tempurl%
\url{https://www.usgs.gov/centers/eros/science/annual-national-land-cover-database}
\showURL{%
\tempurl}


\bibitem[{U.S. Geological Survey}(nd)]%
        {USGSLandsat7}
\bibfield{author}{\bibinfo{person}{{U.S. Geological Survey}}.} \bibinfo{year}{n.d.}\natexlab{}.
\newblock \bibinfo{title}{Landsat 7 Level 2, Collection 2, Tier 1 | Earth Engine Data Catalog}.
\newblock
\urldef\tempurl%
\url{https://developers.google.com/earth-engine/datasets/catalog/LANDSAT_LE07_C02_T1_L2}
\showURL{%
\tempurl}


\bibitem[Zhang and Sabuncu(2018)]%
        {Zhang_Sabuncu_2018}
\bibfield{author}{\bibinfo{person}{Zhilu Zhang} {and} \bibinfo{person}{Mert~R. Sabuncu}.} \bibinfo{year}{2018}\natexlab{}.
\newblock \showarticletitle{Generalized Cross Entropy Loss for Training Deep Neural Networks with Noisy Labels}.
\newblock  \bibinfo{number}{arXiv:1805.07836} (\bibinfo{date}{Nov.} \bibinfo{year}{2018}).
\newblock
\href{https://doi.org/10.48550/arXiv.1805.07836}{doi:\nolinkurl{10.48550/arXiv.1805.07836}}
\newblock
\shownote{arXiv:1805.07836 [cs]}.


\bibitem[Zhi et~al\mbox{.}(2024)]%
        {zhi2024deep}
\bibfield{author}{\bibinfo{person}{Wei Zhi}, \bibinfo{person}{Alison~P Appling}, \bibinfo{person}{Heather~E Golden}, \bibinfo{person}{Joel Podgorski}, {and} \bibinfo{person}{Li Li}.} \bibinfo{year}{2024}\natexlab{}.
\newblock \showarticletitle{Deep learning for water quality}.
\newblock \bibinfo{journal}{\emph{Nature Water}} \bibinfo{volume}{2}, \bibinfo{number}{3} (\bibinfo{year}{2024}), \bibinfo{pages}{228--241}.
\newblock


\end{thebibliography}


\clearpage
\section{Appendix}
\appendix

To help navigate the appendix, Table~\ref{tab:supp_mapping} provides an overview linking each appendix subsection to the corresponding sections in the main paper where they are mentioned or expanded upon.

\begin{table}[h]
  \centering
  \caption{Mapping between appendix sections and corresponding references in the main paper.}
  \renewcommand{\arraystretch}{1.3}
  \setlength{\tabcolsep}{10pt}
  \begin{tabular}{@{}p{0.45\linewidth} p{0.4\linewidth}@{}}
    \toprule
    \textbf{appendix Subsection} & \textbf{Referenced in Main Paper Section(s)} \\
    \midrule
    A. Model Calibration            & Section 5.4 \\
    B. Physically-Informed Pixel Confidence $M_i$ Details          & Section 3.4 \\
    C. Kriging       & Section 4.1 \\
    D. Pollutant Transport Simulation              & Section 4.1 \\
    E. Consistency Analysis    & Section 5.4 \\
    F. Evaluating the Impact of Noise Masks on
Segmentation of Land Cover Types   & Section 4.3 \\
    G. Dataset             & Section 3.1 \\
    H. \pname Per-Class Results    & - \\
    I. Ternary Classification of Hazard Levels        & Section 3.1.1 \\
    J. Statistical Significance of Performance Gains      & - \\
    K. Distance-to-Industry Rasters.           & Section 3.1.2 \\
    L. Ablation Results for Image Resolution    & Section 4.3 \\
    M. Hyperparameter Tuning and Selection      & Section 4.2 \\
    N. Technical Details on Sample Analysis and
Training Data for Real-World Predictions        & Section 5.2 \\
    O. \pname Loss Derivation and Assumptions       & Section 3.5 \\
    P. Region-Level Risk Signals and Bioaccumulation Evidence      & - \\
    \bottomrule
  \end{tabular}
  \label{tab:supp_mapping}
\end{table}

\section{Model Calibration}

We evaluated model calibration using ECE, which quantifies the discrepancy between predicted probabilities and observed outcomes where lower scores indicate better calibration. Table~\ref{tab:ece_scores} shows the ECE scores for our model trained with \pname loss for the years 2008, 2019, and 2022.

\begin{table}[ht!]
  \centering
  \caption{ECE Scores for Models Trained with \pname Loss Across Different Years.}
  \setlength{\tabcolsep}{5pt}  
  \renewcommand{\arraystretch}{1.3}  
  \begin{tabular}{l c c c}
    \toprule
    \textbf{Metric} & \textbf{2008} & \textbf{2019} & \textbf{2022} \\
    \midrule
    ECE Score       & 0.1 & 0.1 & 0.07 \\
    \bottomrule
  \end{tabular}
  \captionsetup{justification=centering, position=bottom}  
  \label{tab:ece_scores} 
\end{table}

The very low ECE values, that is, 0.1 for 2008 and 2019, and 0.07  for 2022, demonstrate that our model’s predicted probabilities are closely aligned with the observed outcomes, indicating excellent calibration.



\section{Physically-Informed Pixel Confidence $M_i$ Details}

we generate a domain expert-informed probability of the label correctness 
for each surface water pixel \(i\):
\begin{align}
    p_{\text{final}} = & \, \alpha_1 \cdot p_{\text{dischargers}} + \alpha_2 \cdot p_{\text{landcover}} + \alpha_3 \cdot p_{\text{sample\_dist}} \nonumber \\
    & \, + \alpha_4 \cdot p_{\text{downstream}}
\end{align}
where \(p_{\text{dischargers}}\),  \(p_{\text{landcover}}\),  \(p_{\text{sample\_dist}}\), and  \(p_{\text{downstream}}\) represent  probabilities derived from (i) proximity to PFAS dischargers, (ii) land cover type, (iii) distance to other sample points, and (iv) downstream flow, respectively. In  detail:

\begin{itemize}
    \item \textbf{Proximity to PFAS Dischargers (\(p_{\text{dischargers}}\))} 
Proximity to known PFAS dischargers is a strong indicator of contamination risk, as shown in scientific analyses~\footnote{\href{https://pubmed.ncbi.nlm.nih.gov/36398312/}{PFAS dischargers}}.
If a pixel’s label is 1, we apply an exponential decay function of distance, assigning higher probabilities to pixels closer to known dischargers. For pixels labeled 0, we invert this logic so that being near a discharger lowers confidence in the 0 label. 

\item\textbf{Land Cover Type (\(p_{\text{landcover}}\))}
For pixels labeled 1, surrounding urban or built-up areas received higher probabilities, consistent with research linking PFAS to industrial zones ~\footnote{\href{https://pubmed.ncbi.nlm.nih.gov/39369000/}{Landcover}}. Conversely, for 0-labeled pixels, surrounding undeveloped or forested land cover increased confidence in low contamination.

\item\textbf{Proximity to Other Sample Points (\(p_{\text{sample\_dist}}\))}
Pixels closer to ground truth sample points 
were assigned higher probabilities using an exponential decay function based on distance to each sample point in the raster. 


\item\textbf{Downstream Flow (\(p_{\text{downstream}}\))}
PFAS may also travel downstream in aquatic systems, making hydrologic flow another critical factor in contamination spread~\footnote{\href{https://pubmed.ncbi.nlm.nih.gov/35339537/}{Downstream Flow}}.
If a pixel lies downstream of a sample point labeled 1, its probability for label 1 increases, and similarly for label 0. This flow direction channel was generated using ArcGIS Pro~\footnote{\href{https://www.esri.com/en-us/arcgis/products/arcgis-pro/overview}{ArcGIS Pro}} based on digital elevation models (DEMs)~\footnote{\href{https://www.earthdata.nasa.gov/topics/land-surface/digital-elevation-terrain-model-dem}{DEM}}. 

\end{itemize}

Full performance results for all configurations are reported in Table~\ref{tab:noise_ablation_full}, demonstrating that our selected weighting scheme most effectively balances the different environmental signals for robust PFAS contamination prediction.

\begin{table}[ht!]
  \centering
  \caption{Performance across all noise mask weight configurations using the 2008 PFAS dataset.}
  \resizebox{\columnwidth}{!}{%
  \begin{tabular}{rrrrrrrrr}
    \toprule
    \textbf{Dischargers (\%)} & \textbf{Landcover (\%)} & \textbf{Flow Dir. (\%)} & \textbf{Sample Dist. (\%)} & \textbf{IoU (\%)} & \textbf{F-score (\%)} & \textbf{Precision (\%)} & \textbf{Recall (\%)} & \textbf{Accuracy (\%)} \\
    \midrule
     \textbf{40} & \textbf{20} & \textbf{30} & \textbf{10} & \textbf{48} & \textbf{61} & \textbf{61} & \textbf{64} & \textbf{73} \\
     40 & 30 & 20 & 10 & 44 & 55 & 58 & 52 & 70 \\
     40 & 10 & 30 & 20 & 46 & 59 & 60 & 56 & 72 \\
     40 & 10 & 20 & 30 & 42 & 53 & 55 & 51 & 69 \\
     40 & 30 & 10 & 20 & 45 & 57 & 59 & 54 & 71 \\
     40 & 20 & 10 & 30 & 43 & 54 & 57 & 52 & 70 \\
     30 & 40 & 10 & 20 & 46 & 58 & 60 & 55 & 72 \\
     30 & 40 & 20 & 10 & 43 & 56 & 57 & 53 & 68 \\
     30 & 10 & 40 & 20 & 44 & 55 & 56 & 53 & 69 \\
     30 & 10 & 20 & 40 & 45 & 57 & 58 & 55 & 70 \\
     30 & 20 & 40 & 40 & 47 & 62 & 60 & 58 & 72 \\
     30 & 20 & 10 & 40 & 45 & 58 & 59 & 56 & 71 \\
     30 & 10 & 30 & 30 & 41 & 52 & 54 & 49 & 69 \\
     10 & 40 & 30 & 20 & 47 & 60 & 60 & 58 & 72 \\
     10 & 40 & 20 & 40 & 44 & 56 & 57 & 54 & 69 \\
     10 & 30 & 40 & 20 & 42 & 53 & 55 & 51 & 70 \\
     10 & 30 & 20 & 40 & 43 & 55 & 56 & 52 & 68 \\
     10 & 20 & 40 & 10 & 46 & 59 & 60 & 57 & 72 \\
     10 & 20 & 30 & 40 & 44 & 56 & 57 & 53 & 70 \\
     20 & 40 & 30 & 40 & 45 & 58 & 59 & 56 & 69 \\
     20 & 40 & 10 & 30 & 46 & 59 & 60 & 57 & 72 \\
     20 & 30 & 40 & 10 & 44 & 57 & 58 & 55 & 69 \\
     20 & 30 & 20 & 30 & 43 & 56 & 57 & 54 & 68 \\
     20 & 10 & 40 & 30 & 40 & 52 & 54 & 50 & 66 \\
    \bottomrule
  \end{tabular}
  }
  \label{tab:noise_ablation_full}
\end{table}

\section{Kriging}
We employed Kriging as a spatial interpolation technique to estimate PFAS. First, we examined the spatial structure of our data by calculating an empirical semivariogram. Using latitude and longitude coordinates alongside the binary target variable, we computed pairwise distances and differences to produce a semivariogram plot, enabling us to observe how variance changed with increasing distance. This step informed the selection of appropriate variogram models (e.g., linear or spherical) for Kriging. Based on this analysis, we selected a spherical variogram model, which best captured the spatial dependencies observed in our data.

Next, we split the data on a state-by-state basis, ensuring geographic diversity between the training and testing subsets. We explored both Ordinary Kriging (assuming a stationary mean across the study area) and Universal Kriging (introducing a drift term to account for regional linear trends), but the two approaches produced comparable results on our dataset. After training each Kriging model on the geographic regions in the training set, we predicted PFAS presence at test set locations. Since we treated PFAS presence as a binary outcome, the resulting continuous Kriging predictions were thresholded at 0.5. We then computed standard classification metrics (such as precision, recall, and F1-score) to evaluate each model’s ability to distinguish between high contamination versus low contamination sites.

While Kriging provided a practical means of interpolating PFAS presence, its reliance on variogram assumptions and spatial stationarity can limit accuracy in areas with highly heterogeneous conditions. 

\section{Pollutant Transport Simulation}
Instead of running a simulation like SWAT directly, which is designed for general watershed hydrology modeling but lacks established parameterization for PFAS transport, we implemented a streamlined Python-based workflow more tailored to PFAS-related surface water contamination. This approach allowed us to integrate domain-relevant features without relying on assumptions unsupported by current PFAS science.

\begin{figure}[ht!]
  \centering
  \includegraphics[width=0.5\columnwidth]{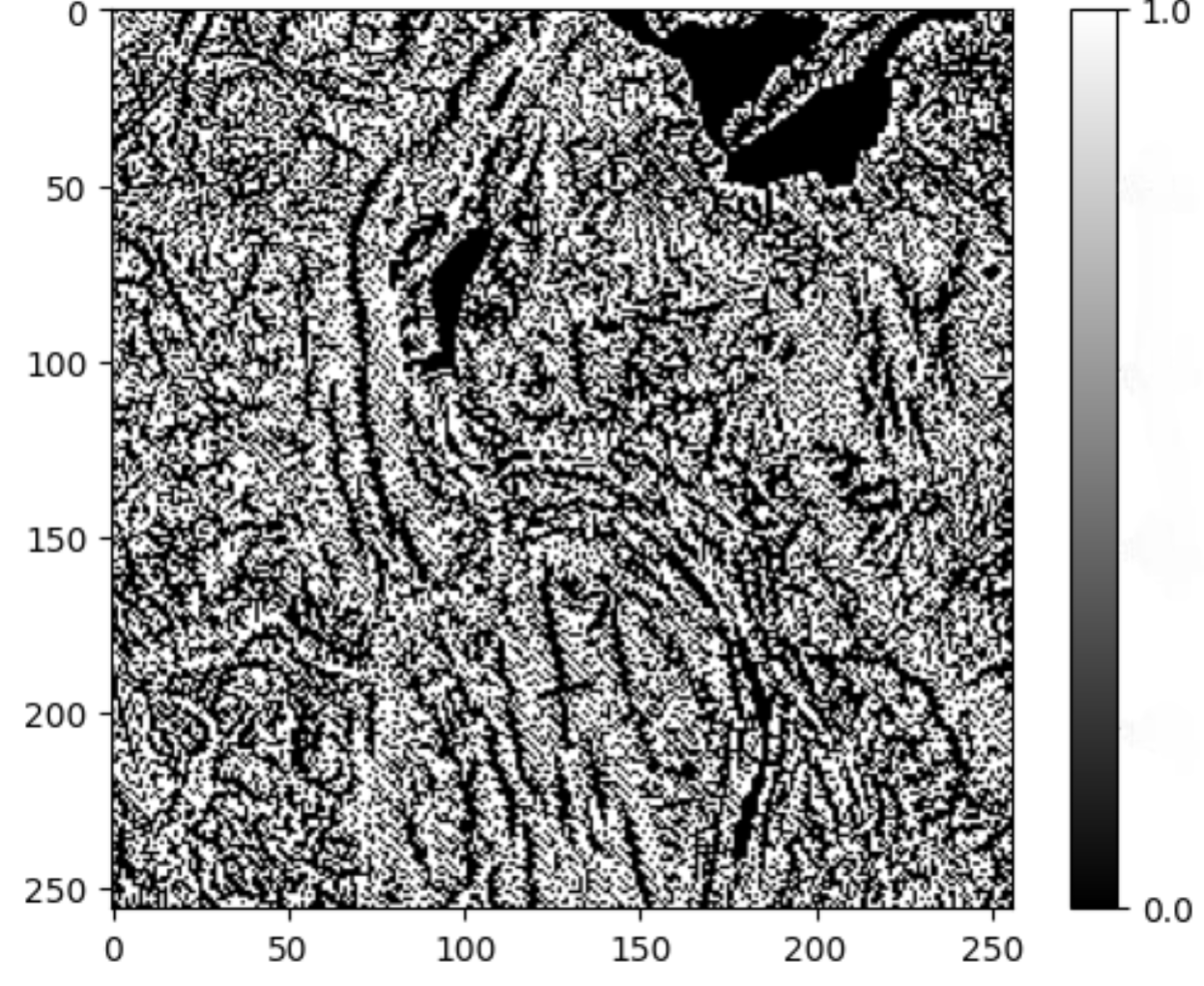}
  \caption{Example binary output from the pollutant transport simulation: 0 represents low contamination, and 1 represents high contamination. The simulation approximates the distribution of PFAS contamination based on hydrological and environmental parameters, providing a practical alternative to full-scale SWAT simulations.}
  \label{fig:simulation_output_example}
\end{figure}

First, we assign initial pollutant concentrations in each patch by setting cells flagged as “dischargers” to a high baseline value (e.g., 100), while land cover–based default concentrations provide moderately-low to low initial contamination levels for other cells. We then construct a Hydrologic Response Unit (HRU) parameter table that assigns infiltration and runoff values based on a cell’s land cover, soil type, and slope. Using these parameters, each patch’s land cover and soil rasters define infiltration/runoff ratios for every grid cell. We incorporate two key rasters; flow direction and flow accumulation, both exported from System for Automated Geoscientific Analyses (SAGA); a GIS and geospatial analysis tool focused on geospatial data processing, analysis, and visualization, to specify how water and pollutant mass move downstream. In each iteration, a fraction of the pollutant mass in each cell is transferred to its downstream neighbor according to the cell’s infiltration/runoff factors, guided by the flow direction raster and scaled by the flow accumulation values. Repeating this process causes pollutant mass to concentrate in lower-lying or higher-flow cells, thereby modeling how contamination evolves over time within the patch.

After the simulation converges, we produce a final pollutant concentration raster, thresholded by the median value of concentrations across all patches for the year to generate a binary contamination map, as illustrated in Fig.~\ref{fig:simulation_output_example}. The binary outputs, with values of 0 and 1 representing low and high contamination respectively, visualizes the simulation's predictions of contamination distribution. These outputs are compared against test set patches containing actual observed PFAS presence \textit{in surface water} to evaluate how accurately the simulation captures observed contamination patterns. Although this standalone approach remains computationally non-trivial, it is more tractable for batch processing of multiple patches than running a full SWAT project repeatedly. Consequently, it provides a practical baseline for pollutant transport modeling within our broader framework, allowing us to compare simulated outputs to both real-world data and other modeling approaches.

\section{Consistency Analysis}
To evaluate local spatial consistency, we applied a robustness test across all 49 continental U.S. states. For each state, we extracted two overlapping image patches; one with a 56×56 pixel overlap and another with 156×156, using three different random locations to ensure stability. We then compared the model’s predictions over the overlapping regions by computing pixel-wise agreement. Agreement is the proportion of matching predicted labels across overlapping pixels between two patches:

\begin{equation}
\text{Agreement} = \frac{1}{N} \sum_{i=1}^{N} \mathbf{1}(y_i^{(1)} = y_i^{(2)}),
\end{equation}

where \( y_i^{(1)} \) and \( y_i^{(2)} \) are the predicted labels for the \(i^{\text{th}}\) pixel in each patch, and \(N\) is the total number of overlapping pixels.

Table~\ref{tab:consistency} reports the agreement across different years using our final \pname model. The high consistency scores across both overlap sizes demonstrate that the model generates stable predictions for the same spatial region even when seen in different contexts. This supports the reliability of our framework in real-world geospatial settings.

\begin{table}[h]
  \centering
  \caption{Spatial consistency measured as prediction agreement over overlapping patch regions.}
  \setlength{\tabcolsep}{6pt}
  \renewcommand{\arraystretch}{1.2}
  \begin{tabular}{lcc}
    \toprule
    \textbf{Year} & \textbf{56×56 Overlap} & \textbf{156×156 Overlap} \\
    \midrule
    2008 & 95 $\pm$ 1 \% & 93 $\pm$ 1 \%  \\
    2019 & 96 $\pm$ 2 \% & 98 $\pm$ 2 \% \\
    2022 & 93 $\pm$ 1 \% & 94 $\pm$ 1 \% \\
    \bottomrule
  \end{tabular}
  \captionsetup{justification=centering}
  \label{tab:consistency}
\end{table}

\section{Evaluating the Impact of Noise Masks on Segmentation of Land Cover Types}

In this experiment, as \textbf{introduced in Section~3.3}, we \pname on distinguishing between two pairs of land cover (LC) types:
\begin{itemize}
    \item \textbf{Evergreen Forests vs Barren Land} (Class 0: Barren Land, Class 1: Evergreen Forest)
    \item \textbf{Deciduous Forests vs Water} (Class 0: Deciduous Forest, Class 1: Water)
\end{itemize}



In this experiment, we simulate uncertainty in the labeling process to align with the uncertainty potentially present in our PFAS ground truth label expansion. Specifically, we randomly throw out half of the valid class pixels and replace them with the other class’s label. 
We then generate noise masks using only the NDVI values, which are used during the training process to handle the uncertainty. 


\noindent\textbf{Results}
Below are the macro-averaged results showing the performance metrics for the Evergreen Forests vs Barren Land and Deciduous Forests vs Water pairs, both with and without noise masks. These LC image patches are aligned with the same spatial footprint as the PFAS patches to ensure consistency in coverage. The evaluation is conducted on at most 2 pixels per image patch, simulating sparse ground truth availability similar to the PFAS setting. We see that including noise masks is useful in this setting as well.

\begin{table}[ht!]
  \centering
  \caption{Performance Comparison for Evergreen Forests vs Barren Land: Simulated Uncertainty in PFAS Pseudolabeling}
    \setlength{\tabcolsep}{5pt}  
    \resizebox{\columnwidth}{!}{  
    \begin{tabular}{l c c c c c}
      \toprule
      \multicolumn{3}{c}{\textbf{Evergreen Forests vs Barren Land}} & \multicolumn{2}{c}{\textbf{Performance Metrics}} \\
      \cmidrule(lr){2-6}
      \textbf{Method} & \textbf{IoU} & \textbf{F1-Score} & \textbf{Precision} & \textbf{Recall} & \textbf{Accuracy} \\
      \midrule
      With Noise Masks  & 0.38 & 0.55 & 0.56 & 0.56 & 0.55 \\
      Without Noise Masks  & 0.24 & 0.33 & 0.74 & 0.50 & 0.48 \\
      \bottomrule
    \end{tabular}}
  \captionsetup{justification=centering, position=bottom}  
  \label{tab:evergreen_barren}
\end{table}

\begin{table}[ht!]
  \centering
  \caption{Performance Comparison for Deciduous Forests vs Water: Simulated Uncertainty in PFAS Pseudolabeling}
    \setlength{\tabcolsep}{5pt}  
   \resizebox{\columnwidth}{!}{ 
    \begin{tabular}{l c c c c c}
      \toprule
      \multicolumn{3}{c}{\textbf{Deciduous Forests vs Water}} & \multicolumn{2}{c}{\textbf{Performance Metrics}} \\
      \cmidrule(lr){2-6}
      \textbf{Method} & \textbf{IoU} & \textbf{F1-Score} & \textbf{Precision} & \textbf{Recall} & \textbf{Accuracy} \\
      \midrule
      With Noise Masks  & 0.38 & 0.55 & 0.55 & 0.56 & 0.55 \\
      Without Noise Masks  & 0.33 & 0.49 & 0.51 & 0.51 & 0.52 \\
      \bottomrule
    \end{tabular}}
  \captionsetup{justification=centering, position=bottom}  
  \label{tab:deciduous_water}
\end{table}

\section{Dataset}
The dataset was curated to ensure that patches included in the training and testing sets were geographically disjoint, avoiding any overlap. 

\begin{figure}[ht!]
  \centering
  \includegraphics[width=0.5\columnwidth]{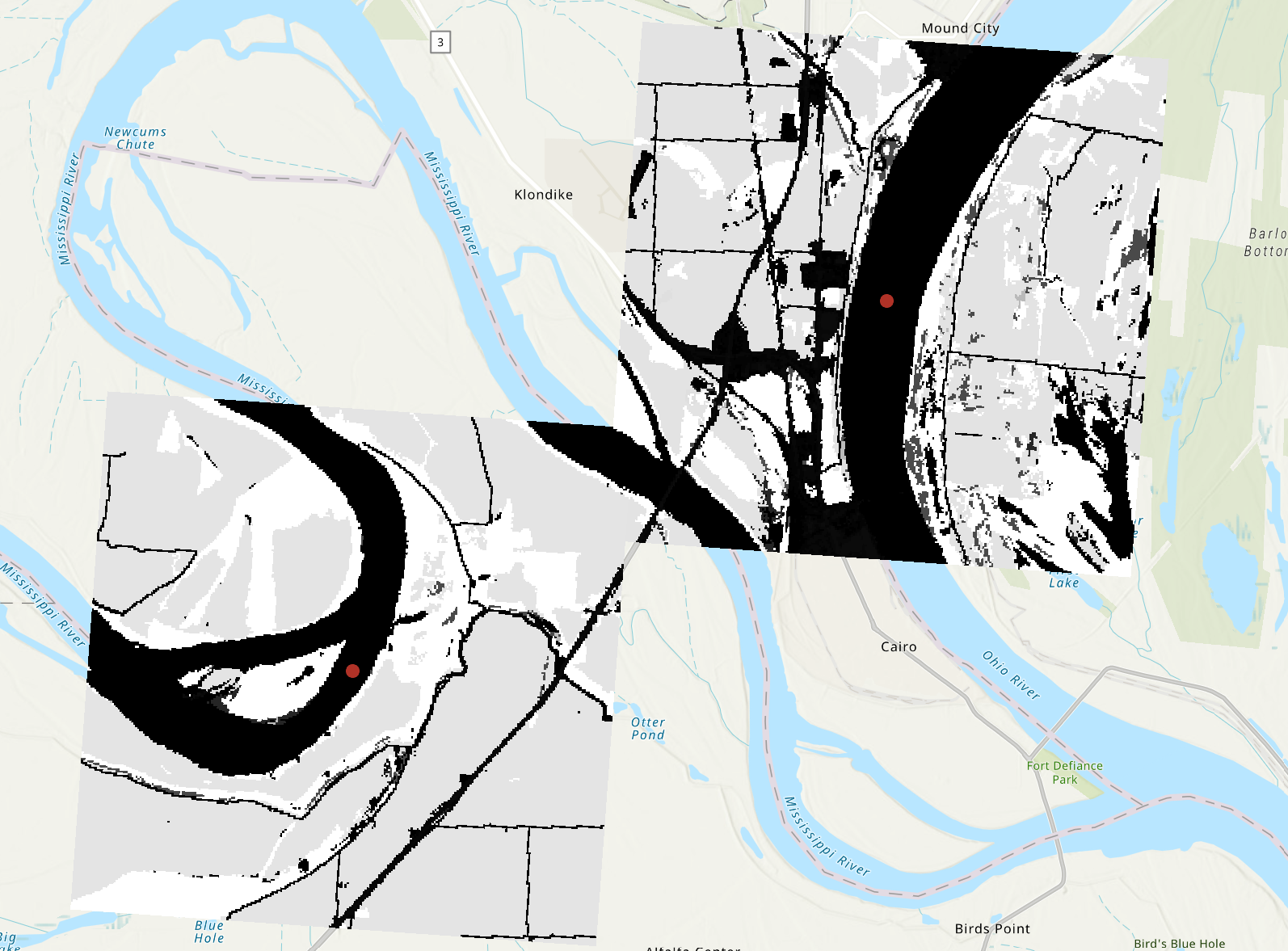}
  \caption{Example of overlapping patch assignment. The figure illustrates two slightly overlapping patches overlaid on the U.S. map, each containing distinct sample points. To maintain disjoint training and testing sets, both patches are assigned to the same set (either training or testing), ensuring no part of a test patch overlaps with any patch seen during training.}
  \label{fig:overlapping_patch_assignment}
\end{figure}

In cases where two or more patches overlapped, as illustrated in Fig.~\ref{fig:overlapping_patch_assignment}, all such patches were assigned to the same set (either training or testing). 
This ensured that no part of a test patch had been seen during training, preserving the integrity of the evaluation.

Additionally, Table \ref{tab:dataset_summaryy} expands on the table in \textbf{Section 3.5.1} to include class distribution. There is a significant class imbalance that we observe in the real world: unfortunately, a majority of water bodies have PFAS contamination that exceed health guidelines, as found in prior scientific work on PFAS prediction as well 
\footnote{See: \textit{PFAS pollute 83\% of U.S. waterways}, E\&E News by Politico (2023). Available at: \url{https://www.eenews.net/articles/pfas-pollute-83-of-u-s-waterways/}}
. To affirm that there is in fact signal to our predictions, we provide class-specific results in Section H. For instance, we find approximately 80\% F-score for class 0 for the year 2024, showing we are truly predicting class 0, and better than random. 
We also add a ternary classification in Section I 
to provide additional levels of risk.

\begin{table}[h]
\centering
\caption{Datasets used for training, testing, and real-world validation, including total size and class balance.}
\resizebox{\columnwidth}{!}{
\begin{tabular}{lllll}
\toprule
\textbf{Type} & \textbf{Use} & \textbf{Data Source} & \textbf{Total Size} & \textbf{\% Pos / Neg} \\
\midrule
Fish  & Train/Test        & EPA \cite{USEPA2015,US_EPA_2024}             & 866 & 89.5\% / 10.5\% \\
Fish  & Validation        & MPART \cite{egle_fcmp_2025}                  & 114 & 77.2\% / 22.8\%  \\
Water & Train/Test        & MPART \cite{egle_pfas_surface_water}         & 293 & 66.9\% / 33.1\% \\
Water & Real-World Validation & Sampled by Authors                      & 8   & 100\% / 0\% \\
\bottomrule
\end{tabular}
}
\label{tab:dataset_summaryy}
\end{table}

\section{\pname Per-Class Results}
To better understand model behavior across classes, Table~\ref{tab:ablation_classwise} reports per-class performance metrics for \pname across each year, including results on the 2024 MPART surface water dataset.

\begin{table}[h]
  \centering
  \caption{Per-class performance of our model using \pname across four years (MPART surface water results at 2024). Results averaged over 3 random seeds.}
    \setlength{\tabcolsep}{5pt}
    \resizebox{\columnwidth}{!}{
    \begin{tabular}{l cc cc cc cc}
      \toprule
      & \multicolumn{2}{c}{\textbf{2008 (\%)}} & \multicolumn{2}{c}{\textbf{2019 (\%)}} & \multicolumn{2}{c}{\textbf{2022 (\%)}} & \multicolumn{2}{c}{\textbf{2024 (\%)}} \\
      \cmidrule(lr){2-3} \cmidrule(lr){4-5} \cmidrule(lr){6-7} \cmidrule(lr){8-9}
      \textbf{Metric} & \textbf{Class 0} & \textbf{Class 1} & \textbf{Class 0} & \textbf{Class 1} & \textbf{Class 0} & \textbf{Class 1} & \textbf{Class 0} & \textbf{Class 1} \\
      \midrule
      Accuracy  & 73 $\pm$ 1 & 73 $\pm$ 1 & 82 $\pm$ 1 & 82 $\pm$ 1 & 88 $\pm$ 2 & 88 $\pm$ 2 & 79 $\pm$ 2 & 79 $\pm$ 1 \\
      IoU       & 25 $\pm$ 2 & 70 $\pm$ 2 & 40 $\pm$ 2 & 79 $\pm$ 2 & 47 $\pm$ 3 & 87 $\pm$ 2 & 67 $\pm$ 2 & 63 $\pm$ 2 \\
      F-score   & 40 $\pm$ 2 & 82 $\pm$ 2 & 57 $\pm$ 2 & 88 $\pm$ 2 & 64 $\pm$ 3 & 93 $\pm$ 2 & 80 $\pm$ 2 & 77 $\pm$ 2 \\
      Precision & 33 $\pm$ 3 & 88 $\pm$ 2 & 42 $\pm$ 2 & 98 $\pm$ 2 & 60 $\pm$ 3 & 94 $\pm$ 1 & 75 $\pm$ 1 & 83 $\pm$ 2 \\
      Recall    & 50 $\pm$ 2 & 78 $\pm$ 2 & 89 $\pm$ 1 & 80 $\pm$ 1 & 69 $\pm$ 1 & 92 $\pm$ 2 & 86 $\pm$ 1 & 71 $\pm$ 1 \\
      \bottomrule
    \end{tabular}}
  \label{tab:ablation_classwise}
\end{table}

\section{Ternary Classification of Hazard Levels}\label{sec:3class}

To enable finer-grained prioritization of contamination severity, we extended our binary classification framework to a ternary setting. Specifically, each fish tissue sample was assigned to one of three classes based on its hazard index (HI), defined as the ratio of measured PFAS concentration to the EPA’s reference dose:

\begin{itemize}
    \item \textbf{Class 0 (Low):} HI $<$ 1 (i.e., below the EPA baseline threshold)
    \item \textbf{Class 1 (Medium):} $1 \leq$ HI $\leq 1000$
    \item \textbf{Class 2 (High):} HI $>$ 1000
\end{itemize}

The threshold of 1000 was informed by a combination of data-driven distributional patterns and recent regulatory guidance. In particular, the updated 2025 Michigan Department of Health and Human Services fish consumption guidelines define a 50~ppb (do-not-eat) threshold; approximately a 900$\times$ range relative to the EPA’s 0.056~ppb PFOS reference dose. While these thresholds may vary by state, the Michigan values are among the most recent and provide a concrete, health-based rationale for distinguishing especially high-risk cases. It also corresponds well with natural breakpoints observed in our hazard index distributions. 

\begin{table}[ht!]
  \centering
  \caption{Performance of \pname under ternary classification (0: low, 1: medium, 2: high). Results averaged over 3 random seeds.}
    \setlength{\tabcolsep}{6pt}
    \renewcommand{\arraystretch}{1.3}
    \begin{tabular}{l c c c}
      \toprule
      \textbf{Metric} & \textbf{2008 (\%)} & \textbf{2019 (\%)} & \textbf{2022 (\%)} \\
      \midrule
      Accuracy  & 55 $\pm$ 1 & 72 $\pm$ 1 & 65 $\pm$ 1 \\
      IoU       & 38 $\pm$ 1 & 48 $\pm$ 2 & 44 $\pm$ 2 \\
      F-score   & 54 $\pm$ 1 & 63 $\pm$ 1 & 61 $\pm$ 1 \\
      Precision & 66 $\pm$ 2 & 70 $\pm$ 2 & 60 $\pm$ 2 \\
      Recall    & 65 $\pm$ 2 & 61 $\pm$ 1          & \textbf{62 $\pm$ 2} \\
      \bottomrule
    \end{tabular}
  \label{tab:ablation_noise_ternary_FOCUS_only}
\end{table}

Results demonstrate signal, but are less strong compared to \pname binary classification performance, given the added complexity of the ternary task. 

\section{Statistical Significance of Performance Gains}

To evaluate the robustness of our improvements over the focal-only baseline, we conducted statistical significance testing using the Wilcoxon signed-rank test, a non-parametric paired test appropriate for comparing performance across random seeds.

For each year (2008, 2019, 2022), we collected the F-scores from multiple runs (5 seeds per year) and compared the \pname results against focal-only using a two-sided Wilcoxon signed-rank test. As shown in Table~\ref{tab:wilcoxon}, \pname achieved statistically significant improvements in all three years ($p = 0.03125$), despite the relatively small number of seeds. Notably, for every random seed, the F-score achieved with \pname was consistently higher than the focal-only counterpart.

These results confirm that the performance gains from our noise-aware loss are not due to chance, but are statistically reliable across different random initializations.

\begin{table}[ht!]
  \centering
  \caption{Wilcoxon signed-rank test comparing F1-scores of \pname and focal-only across 5 random seeds for each year.}
  \setlength{\tabcolsep}{12pt}
  \renewcommand{\arraystretch}{1.3}
  \begin{tabular}{lcc}
    \toprule
    \textbf{Year} & \textbf{Wilcoxon Statistic} & \textbf{$p$-value} \\
    \midrule
    2008 & 0.0 & 0.03125 \\
    2019 & 0.0 & 0.03125 \\
    2022 & 0.0 & 0.03125 \\
    \bottomrule
  \end{tabular}
  \label{tab:wilcoxon}
\end{table}

\section{Distance-to-Industry Rasters}
To capture potential industrial sources of PFAS contamination, we constructed distance rasters for each relevant facility type using data from the EPA's ECHO database. For each year (e.g., 2008, 2019, 2022), we identified all facilities that were active \textit{on or before} that year, ensuring historical context was preserved. Facility types were grouped by industrial category (e.g., airports, landfills, chemical manufacturers), and a separate Euclidean distance raster was generated for each category, where pixel values represent the distance to the nearest facility of that type.

The total number of distance raster channels therefore varies by year, depending on which facility types were present in the ECHO dataset up to that point. For example, a given year may include distance rasters for 15 industry types, while another may include 41.  

\begin{figure}[ht!]
  \centering
  \includegraphics[width=0.7\columnwidth]{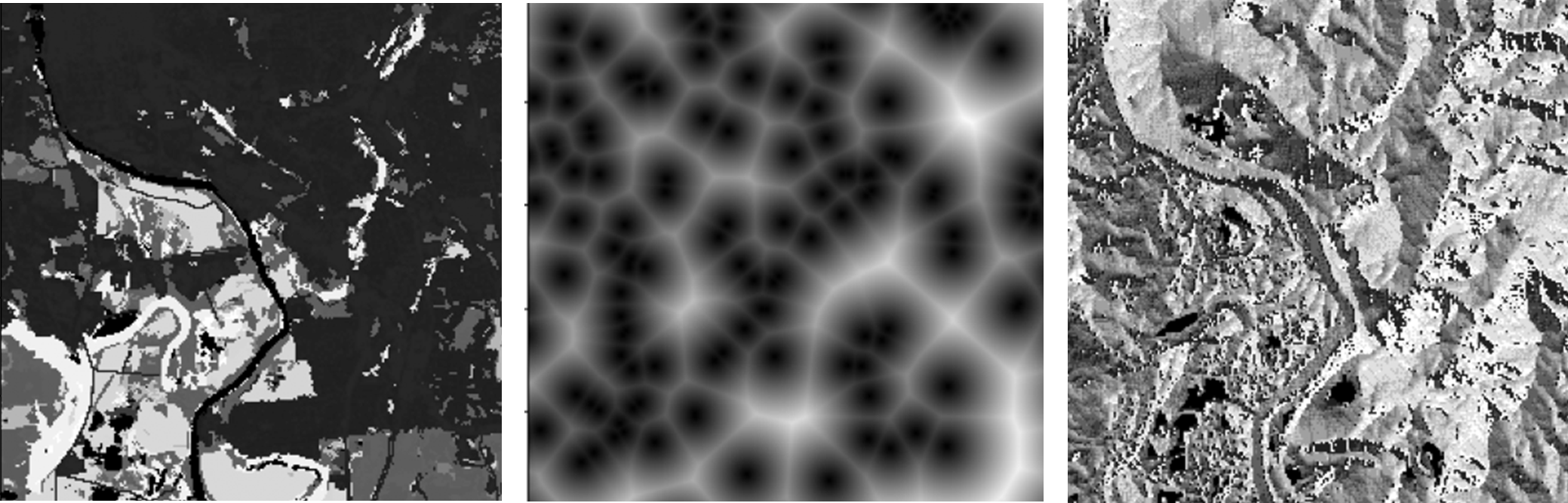}
  \caption{Example raster channels in an image: (left) land cover, (center) distances from  
  chemical manufacturing industries, and (right)  flow direction. }
  \label{fig:test1}
\end{figure}

\section{Ablation Results for Image Resolution}

Table~\ref{tab:ablation_resolution} presents the results of our image resolution ablation study. We compare the model's performance when trained on \(256 \times 256\) versus \(512 \times 512\) input resolutions, without the use of noise-aware training.

\begin{table}[h]
  \centering
  \caption{Comparison of performance between \(256 \times 256\) and \(512 \times 512\) image resolutions without noise-aware training. Bold values indicate the better result.}
    \setlength{\tabcolsep}{5pt}
    \renewcommand{\arraystretch}{1.3}
    \begin{tabular}{l c c c c c c}
      \toprule
      & \multicolumn{2}{c}{\textbf{2008 (\%)}} & \multicolumn{2}{c}{\textbf{2019 (\%)}} & \multicolumn{2}{c}{\textbf{2022 (\%)}} \\
      \cmidrule(lr){2-3} \cmidrule(lr){4-5} \cmidrule(lr){6-7}
      \textbf{Metric} & \textbf{256} & \textbf{512} & \textbf{256} & \textbf{512} & \textbf{256} & \textbf{512} \\
      \midrule
      Accuracy  & \textbf{37 $\pm$ 2} & 35 $\pm$ 3 & \textbf{62 $\pm$ 1} & 59 $\pm$ 2 & \textbf{55 $\pm$ 3} & 49 $\pm$ 3 \\
      IoU       & \textbf{22 $\pm$ 3} & 20 $\pm$ 2 & \textbf{41 $\pm$ 2} & 40 $\pm$ 2 & \textbf{36 $\pm$ 3} & 32 $\pm$ 2 \\
      F-score   & \textbf{36 $\pm$ 3} & 32 $\pm$ 3 & \textbf{57 $\pm$ 1} & 54 $\pm$ 2 & \textbf{53 $\pm$ 3} & 50 $\pm$ 3 \\
      Precision & \textbf{54 $\pm$ 2} & 50 $\pm$ 3 & \textbf{63 $\pm$ 2} & 57 $\pm$ 2 & \textbf{63 $\pm$ 3} & 61 $\pm$ 3 \\
      Recall    & \textbf{55 $\pm$ 2} & 51 $\pm$ 2 & \textbf{78 $\pm$ 1} & 70 $\pm$ 2 & \textbf{74 $\pm$ 1} & 70 $\pm$ 2 \\
      \bottomrule
    \end{tabular}
  \label{tab:ablation_resolution}
\end{table}

\section{Hyperparameter Tuning and Selection}

We tuned the learning rate and batch size empirically by monitoring training stability. Specifically, we experimented with learning rates in $\{1 \times 10^{-3}, 5 \times 10^{-4}, 1 \times 10^{-4}\}$ and batch sizes in $\{2, 4, 8\}$. We selected a batch size of 4 because it provided the best stability while preserving the benefits of our pretrained backbone, minimizing the risk of large, destabilizing updates. Notably, these final hyperparameter values are consistent with those adopted in the original Prithvi paper, further supporting their suitability in this context.

\section{Technical Details on Sample Analysis and Training Data for Real-World Predictions}

This section provides additional technical details on the real-world PFAS samples collected and analyzed by our team.

\noindent\textbf{Sample Collection.} New 250~mL HDPE (High-Density Polyethylene) bottles were triple rinsed with ultrapure water before sample collection in the field.

\medskip
\noindent\textbf{Sample Preparation.} Samples were extracted using solid phase extraction, guided by EPA Method 1633. Each sample was spiked with 200~\textmu L of isotopically mass-labeled standards (Wellington).

\medskip
\noindent\textbf{Targeted Sample Analysis.} Targeted PFAS analysis was conducted using a Thermo Exploris 240 Orbitrap mass spectrometer (MS) coupled with a Vanquish ultra-high-performance liquid chromatograph (UHPLC-MS). A C-18 column was used, and the MS was operated in negative electrospray ionization (ESI-) mode.

\medskip
\noindent\textbf{Model Used for Prediction.} The model used to generate predictions for these newly collected samples was trained on 2024 surface water samples from the Michigan MPART dataset. Given the temporal proximity and coverage of this dataset, it provided a decent foundation for making informed predictions on samples collected in mid-2025.

\section{\pname Loss Derivation and Assumptions}
\label{app:focus_derivation}

\noindent\textbf{Setup (latent asymmetric flip noise).}
Let $z_i\in\{0,1\}$ be a latent clean label and $y_i\in\{0,1\}$ the observed noisy label.
Assume per-pixel flip noise with rate $\eta_i \in [0,\tfrac{1}{2})$ (can be asymmetric and spatially varying):
\[
\Pr(y_i = z_i \mid x_i) = 1-\eta_i,
\qquad
\Pr(y_i \neq z_i \mid x_i) = \eta_i.
\]
Let $\pi_\theta(x_i)=\Pr_\theta(z_i=1\mid x_i)$ and define
$\pi_\theta^{(y_i)}(x_i)=\pi_\theta(x_i)$ if $y_i=1$ and $1-\pi_\theta(x_i)$ if $y_i=0$.
Then the noisy likelihood is
\begin{equation}
\label{eq:noisy_like}
\Pr_\theta(y_i\mid x_i)=\eta_i + (1-2\eta_i)\,\pi_\theta^{(y_i)}(x_i),
\end{equation}
and the per-example noisy negative log-likelihood (NLL) is
\begin{equation}
\label{eq:noisy_nll}
\ell_{\mathrm{noisy}}(\theta;x_i,y_i)
= -\log\!\Big(\eta_i + (1-2\eta_i)\,\pi_\theta^{(y_i)}(x_i)\Big).
\end{equation}


\begin{lemma}[Local weighted-CE upper bound]
\label{lem:ce_bound}
Fix $\eta_i\in[0,\tfrac{1}{2})$ and an anchor $p_{0i}\in(0,1)$.
Let $g_i(p)=-\log(\eta_i+(1-2\eta_i)p)$.
Then for all $p$ in a neighborhood of $p_{0i}$,
\begin{equation}
\label{eq:tangent_bound}
g_i(p)\;\le\; g_i(p_{0i}) + g_i'(p_{0i})(p-p_{0i})
\;=\; -w_i\log p + C_i,
\end{equation}
where
\begin{equation}
\label{eq:wi_def}
w_i
=\frac{(1-2\eta_i)p_{0i}}{\eta_i+(1-2\eta_i)p_{0i}}
\in[0,1],
\qquad
C_i = g_i(p_{0i}) + w_i\log p_{0i}.
\end{equation}
\end{lemma}

\noindent\textbf{Interpretation.}
Lemma~\ref{lem:ce_bound} implies the noisy NLL behaves locally like a
\emph{weighted cross-entropy} term $-w_i\log p$, and the weight $w_i$ decreases as
$\eta_i$ increases (labels become less informative).

\noindent\textbf{Assumption on $M_i$ (role of the environmental priors-informed confidence).}
We use $M_i\in[0,1]$ as a practical surrogate for the unknown reliability weight.
Concretely, we assume $M_i$ is \emph{aligned with label cleanliness}:
\noindent\textbf{(Alignment assumption).}
We assume that $\mathbb{E}[M_i \mid \eta_i]$ is non-increasing in $\eta_i$, and that
$M_i$ approximates $w_i$ up to a monotone re-scaling:
\begin{equation}
\label{eq:Mi_assump}
\mathbb{E}[M_i \mid \eta_i] \downarrow \text{ in } \eta_i,
\qquad
M_i \approx w_i .
\end{equation}

Intuitively, $M_i$ is larger near locations where supervision is more trustworthy
(e.g., near sampling points or along plausible hydrologic pathways) and smaller in
regions where point-to-pixel propagation is likely wrong.

We next show that multiplying by a bounded focal factor preserves a local surrogate form.

\begin{lemma}[Focal modulation preserves local surrogate form]
\label{lem:focal_bound}
Fix $\gamma\ge 0$ and anchor $p_{0i}\in(0,1)$.
There exist constants $A_i>0$ and $B_i$ such that for all $p$ near $p_{0i}$,
\begin{equation}
\label{eq:focal_bound}
-\log(\eta_i+(1-2\eta_i)p)\;\le\;
A_i\,(1-p)^\gamma(-\log p) + B_i.
\end{equation}
\end{lemma}

\noindent\textbf{Interpretation.}
Near a fixed operating point $p_{0i}$, $(1-p)^\gamma$ is a positive bounded scalar,
so focal loss is a locally rescaled cross-entropy and retains classification-calibrated behavior.

We now connect the above lemmas to the proposed objective.

\begin{theorem}[\pname as a noisy-NLL surrogate (local)]
\label{thm:focus_surrogate}
Under the model in~\eqref{eq:noisy_like}--\eqref{eq:noisy_nll} and $\gamma\ge 0$,
for each example $i$ there exist constants $\tilde{A}_i>0,\tilde{B}_i$ such that
for all $p$ near $p_{0i}$,
\[
\ell_{\mathrm{noisy}}(\theta;x_i,y_i)
\;\le\;
\tilde{A}_i\, M_i\,(1-p)^\gamma(-\log p) + \tilde{B}_i,
\quad\text{where } p=\pi_\theta^{(y_i)}(x_i).
\]
Equivalently, minimizing
\begin{equation}
\label{eq:focus_app}
\mathcal{L}_{\text{\pname}}(x_i,y_i)
=
M_i\,(1-\pi_\theta^{(y_i)}(x_i))^\gamma\,
\big(-\log \pi_\theta^{(y_i)}(x_i)\big)
\end{equation}
minimizes a valid local upper bound on $\ell_{\mathrm{noisy}}$ (up to constants),
and thus maximizes a corresponding local lower bound on the noisy log-likelihood.
\end{theorem}

\noindent\textbf{Connection to expected noisy risk and calibration.}
Aggregating Theorem~\ref{thm:focus_surrogate} over pixels yields a surrogate for the
expected noisy risk $\mathbb{E}[\ell_{\mathrm{noisy}}]$.
Since the modulating factor is non-negative and bounded for $p\in(0,1)$ and $\gamma\ge 0$,
the objective remains classification-calibrated in the standard sense (locally equivalent to CE).

\noindent\textbf{Edge cases and sanity checks.}
\begin{itemize}
  \item \textbf{Clean-label limit.} If $\eta_i\to 0$, then $w_i\to 1$ in~\eqref{eq:wi_def}.
  If additionally $M_i\to 1$, then \pname reduces to standard focal loss; with $\gamma=0$ it reduces to CE.
  \item \textbf{Uninformative-label limit.} If $\eta_i\to \tfrac{1}{2}$, then $w_i\to 0$.
  Under the alignment assumption~\eqref{eq:Mi_assump}, $M_i$ also becomes small, so \pname naturally down-weights (or effectively ignores) pixels whose supervision is uninformative.
  \item \textbf{Calibration of $M_i$.} In practice $M_i$ need not equal the true $w_i$; it suffices that
  $M_i$ ranks pixels by reliability (monotone alignment). This is precisely the role of our environmental priors-informed construction.
\end{itemize}

\section{Region-Level Risk Signals and Bioaccumulation Evidence}
\label{sec:risk_signals}
\begin{figure}[ht]
\centering
\fbox{%
\includegraphics[width=0.80\columnwidth]{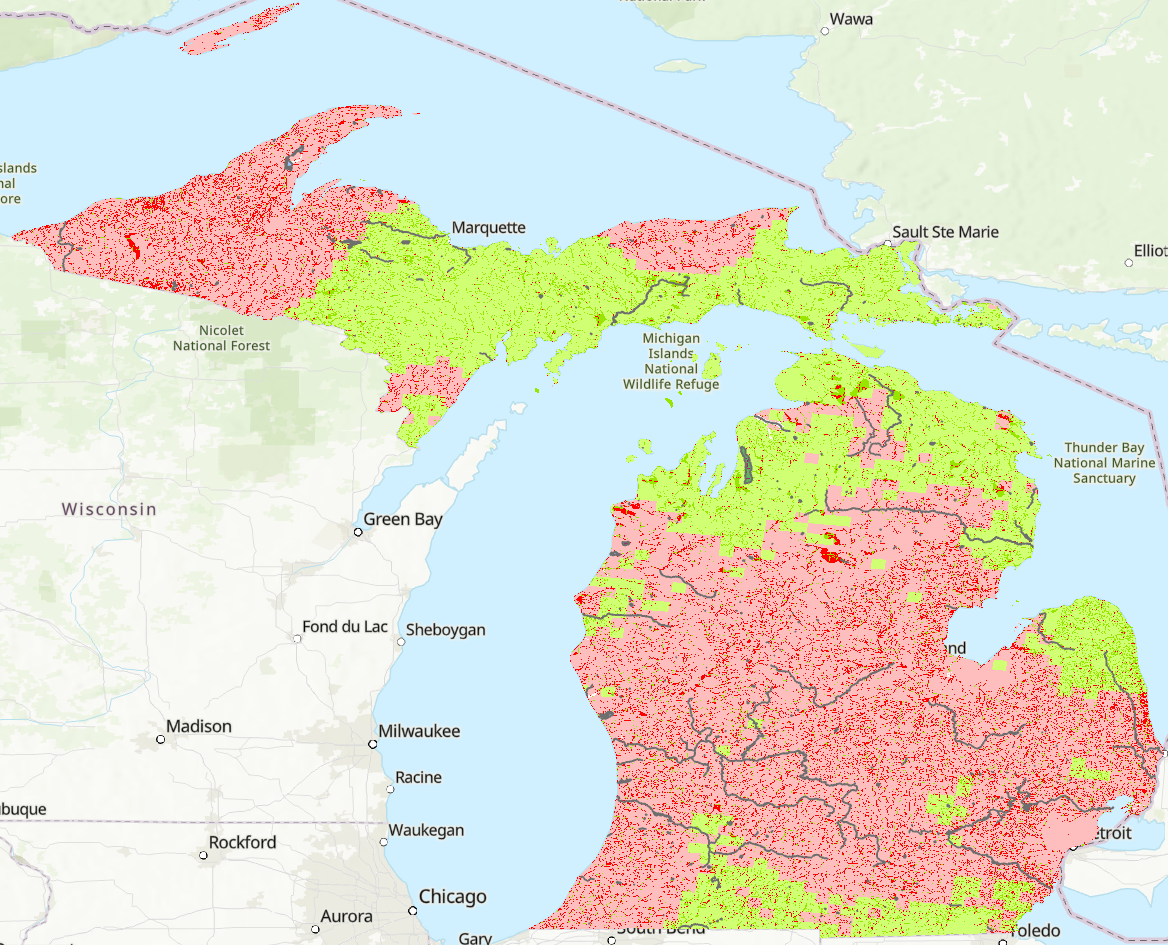}
}
\caption{Region-level aggregation of predicted surface-water PFAS risk over Michigan. Colors indicate low (green), and high (red) predicted risk. Overlaid regions (polylines) correspond to independent fish consumption advisory areas from state monitoring programs.}
\label{fig:risk_placeholder}
\Description{Region-level visualization comparing aggregated surface-water contamination risk with fish advisory evidence.}
\end{figure}

To assess the scientific relevance of model predictions beyond pixel-level accuracy, we aggregate surface-water contamination predictions into region-level risk categories and examine their spatial correspondence with independent fish advisory evidence \cite{MDHHS_FindYourArea_SafeFish} over Michigan (Fig~\ref{fig:risk_placeholder}). Predicted surface-water contamination categories show spatial agreement with regions exhibiting elevated fish contamination, supporting a link between modeled water contamination risk and bioaccumulation-related risk signals.

At the same time, discrepancies between predicted high-risk waters and advisories highlight regions for targeted sampling and scientific follow-up. This analysis illustrates how noise-aware geospatial learning supports AI for Science by integrating large-scale data with sparse observations to guide future measurement efforts.

\end{document}